\documentclass{article} % For LaTeX2e
\usepackage{iclr2025_conference,times}

\usepackage{amsmath,amsfonts,bm}

\def\eqref#1{equation~\ref{#1}}
\def\1{\bm{1}}

\DeclareMathAlphabet{\mathsfit}{\encodingdefault}{\sfdefault}{m}{sl}
\SetMathAlphabet{\mathsfit}{bold}{\encodingdefault}{\sfdefault}{bx}{n}

\usepackage{hyperref}
\usepackage{url}
\usepackage{wrapfig}

\usepackage{hyperref}
\usepackage{url}

\usepackage{algorithm}
\usepackage{algorithmic}
\usepackage{xcolor}

\usepackage{algorithm}
\usepackage{algorithmic}
\usepackage{amsmath}
\usepackage{amssymb}
\usepackage{mathtools}
\usepackage{amsthm}
\usepackage{bm}
\usepackage{enumitem}
\usepackage{algorithm}
\usepackage{algorithmic}
\usepackage{mathtools}

\usepackage{microtype}
\usepackage{graphicx}
\usepackage{multirow}
\usepackage{multicol}
\usepackage{booktabs} % for professional tables
\usepackage{subcaption}
\usepackage{graphicx}

\usepackage{hyperref}
\usepackage{url}
\usepackage{booktabs}
\usepackage{multirow}
\usepackage{graphicx}

\usepackage{amsmath}
\usepackage{amssymb}
\usepackage{mathtools}
\usepackage{amsthm}
\usepackage{bm}
\usepackage{enumitem}
\usepackage{algorithm}
\usepackage{algorithmic}
\usepackage{mathtools}

\usepackage{multirow}
\usepackage[figuresright]{rotating}
\definecolor{mygray}{gray}{.9}
\usepackage{colortbl}

\def\method{\textsc{SMI-Editor}}
\def\mlmmth{\textsc{SMI-MLM}}

\title{\method{}: Edit-based SMILES Language Model with Fragment-level Supervision}

\author{Kangjie Zheng$^{1,2,3}$\!
Siyue Liang$^{1,2,3}$ 
Junwei Yang$^{1,2,3}$ 
Bin Feng$^{4,1,2,3}$
Zequn Liu$^{1,2,3}$ \\
\textbf{Wei Ju}$^{*5}$ 
\textbf{Zhiping Xiao}$^{*6}$ 
\textbf{Ming Zhang}\thanks{Corresponding Authors.}~~\!$^{1,2,3}$ \\
$^1$School of Computer Science, Peking University.\\
$^2$National Key Laboratory for Multimedia Information Processing,  Peking University.\\
$^3$Peking University-Anker Embodied AI Lab, Peking University.\\
$^4$International Digital Economy Academy (IDEA), Shenzhen, China.\\
$^5$College of Computer Science, Sichuan University, Chengdu, China. \\
$^6$Computer Science and Engineering Department, University of Washington, U.S.A.\\
{$\mathtt{juwei@scu.edu.cn,~\  patxiao@uw.edu,~\ mzhang\_cs@pku.edu.cn}$}\\
}

\iclrfinalcopy % Uncomment for camera-ready version, but NOT for submission.
\begin{document}

\maketitle

\begin{abstract}
SMILES, a crucial textual representation of molecular structures, has garnered significant attention as a foundation for pre-trained language models (LMs). However, most existing pre-trained SMILES LMs focus solely on the single-token level supervision during pre-training, failing to fully leverage the substructural information of molecules. This limitation makes the pre-training task overly simplistic, preventing the models from capturing richer molecular semantic information. Moreover, during pre-training, these SMILES LMs only process corrupted SMILES inputs, never encountering any valid SMILES, which leads to a train-inference mismatch. To address these challenges, we propose \method{}, a novel edit-based pre-trained SMILES LM.
\method{} disrupts substructures within a molecule at random and feeds the resulting SMILES back into the model, which then attempts to restore the original SMILES through an editing process.  
This approach not only introduces fragment-level training signals, but also enables the use of valid SMILES as inputs, allowing the model to learn how to reconstruct complete molecules from these incomplete structures. 
As a result, the model demonstrates improved scalability and an enhanced ability to capture fragment-level molecular information. Experimental results show that \method{} achieves state-of-the-art performance across multiple downstream molecular tasks, and even outperforming several 3D molecular representation models. \footnote{Code is released at \href{https://github.com/zhengkangjie/smi-editor}{https://github.com/zhengkangjie/smi-editor}}
\end{abstract}

\section{Introduction}

With the widespread application of AI in molecular-related tasks, enhancing the modeling of SMILES data has become a key research focus. The textual nature of SMILES data makes it possible for us to leverage past experiences from text modeling to address challenges in SMILES representation. Additionally, the knowledge extracted from SMILES data often aligns well with textual knowledge. A large number of studies have attempted to design SMILES language models (LMs) to explore the knowledge inherent in SMILES sequences \citep{wang2019smiles,chithrananda2020chemberta,bagal2021molgpt,ross2022large}. Significant efforts have also been made to align learned knowledge from SMILES with textual knowledge \citet{edwards2022translation,pei2023biot5,liu2023molxpt}, aiming at boosting the performance of downstream applications such as property prediction and molecular design. A critical issue in these model designs lies in how to efficiently mine molecular-related knowledge from SMILES data. This paper address this issue by presenting a SMILES LM with enhanced modeling capabilities.

Current SMILES LMs often adopt strategies used in natural language processing, such as predicting missing tokens in corrupted SMILES sequences (e.g., Masked Language Modeling (MLM) and Causal Language Modeling (CLM)). However, this approach introduce several challenges:
\begin{enumerate}
    \item [(i)] Unlike natural language, where individual tokens represent independent semantic units like subwords, words or phrases, SMILES data tokens correspond to single atoms, chemical bonds, or special symbols. Molecular functionality, however, is more closely tied to specific substructures (e.g., functional groups). The SMILES LMs focusing only on individual tokens in the SMILES context may fail to capture the semantic information of these substructures. 
    \item[(ii)] Predicting a single missing token in a given SMILES context is often trivial, leading to rapid saturation of model capacity during training. This limits the model's ability to acquire more nuanced molecular knowledge, and affects its scalability and its effectiveness in generalizing to diverse molecular data.
    \item[(iii)] Since these models are trained on corrupted SMILES sequences containing special symbols like \texttt{[MASK]} -- which do not appear in real-world SMILES data -- they face challenges in modeling the semantic content of complete SMILES sequences.
\end{enumerate}

To address these challenges, we propose an edit-based SMILES language model with fragment-level supervision:
\begin{enumerate}
    \item [(i)] To enable the model to learn richer substructure-related molecular information, we introduce fragment-level supervision. By randomly removing substructures from molecules and train the model to recover the missing information, we encourage it to acquire more comprehensive fragment-level semantic knowledge.
    \item[(ii)] We design an edit-based pre-training objective, allowing the model to process valid SMILES sequences and to restore missing substructures through an editing process.
\end{enumerate}

In summary, our contributions are threefold:

\begin{itemize}[leftmargin=*]
    \item We analyze the behavior of current SMILES masked language models (MLMs) during pre-training phase and downstream tasks, identifying their rapid saturation problem in pre-training and their limited ability to capture the molecular substructure information. Previous studies lack a systematic exploration of these issues.
    \item We introduce the first edit-based pre-trained LM for SMILES, which transforms valid SMILES sequences into structurally related variants. This approach resolves the train-inference mismatch issue in existing SMILES LMs. The integration of fragment-level supervision further enhances our model's ability to learn richer semantic information from SMILES and improves its overall performance.
    \item Extensive experiments demonstrate that \method{} achieves state-of-the-art performance across multiple molecular property prediction tasks, surpassing several 3D molecular representation models. Ablation studies confirm the effectiveness and scalability of \method{}.

\end{itemize}

\section{Understanding the Behavior of MLM}
\label{others}

\begin{wrapfigure}{r}{0.25\textwidth}
\vspace{-2em}
\begin{center}
\includegraphics[width=\linewidth]{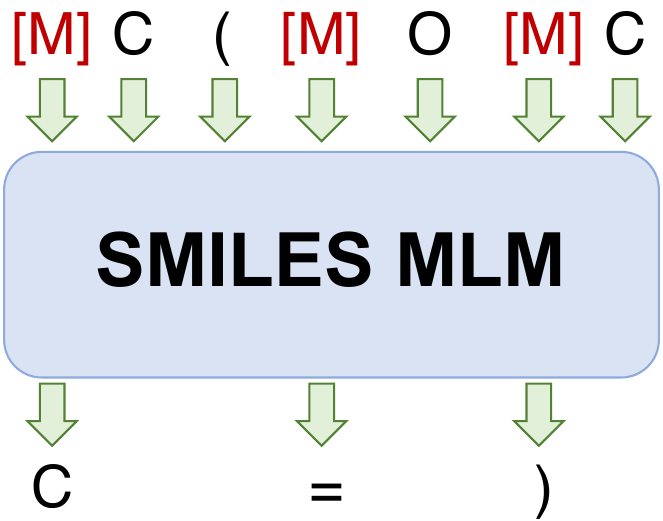}
\end{center}
\vspace{-1.5em}
\caption{The framework of MLM for SMILES.}

\label{fig:smiles_MLM}
\end{wrapfigure}

The Masked Language Model (MLM) is a widely used approach for textual information modeling and has been extensively applied to SMILES representation learning \citep{wang2019smiles,chithrananda2020chemberta,ross2022large}. During the training process, tokens in a SMILES sequence -- representing single atoms, chemical bonds, or special symbols -- are randomly masked with a fixed masked ratio of 15\%. The model is then trained to predict these masked tokens accurately, as is illustrated in Figure 
\ref{fig:smiles_MLM}. To evaluate the effectiveness and capabilities of MLMs for SMILES data, we conducted a comprehensive series of experiments. 

\begin{figure*}[t]    
% \vspace{-0.4cm} 
  \centering      
  \subfloat[Training Curves: Model Scales]  
  {
      \label{fig:sat_subfig1}\includegraphics[width=0.32\linewidth]{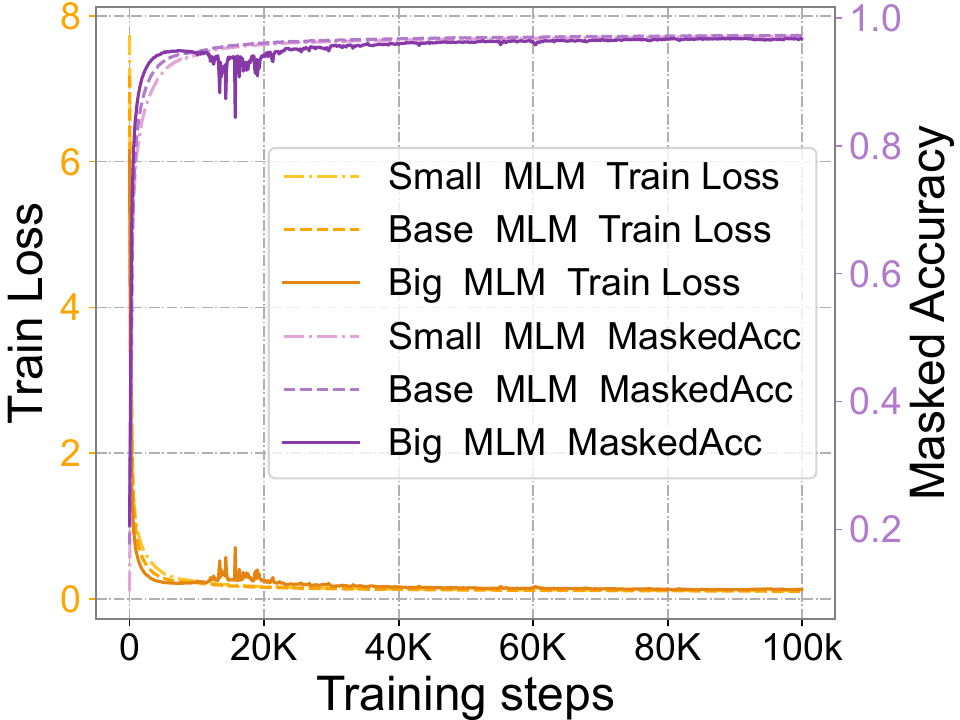}
  }
  \subfloat[Validation Curves: Model Scales]
  {
      \label{fig:sat_subfig2}\includegraphics[width=0.32\linewidth]{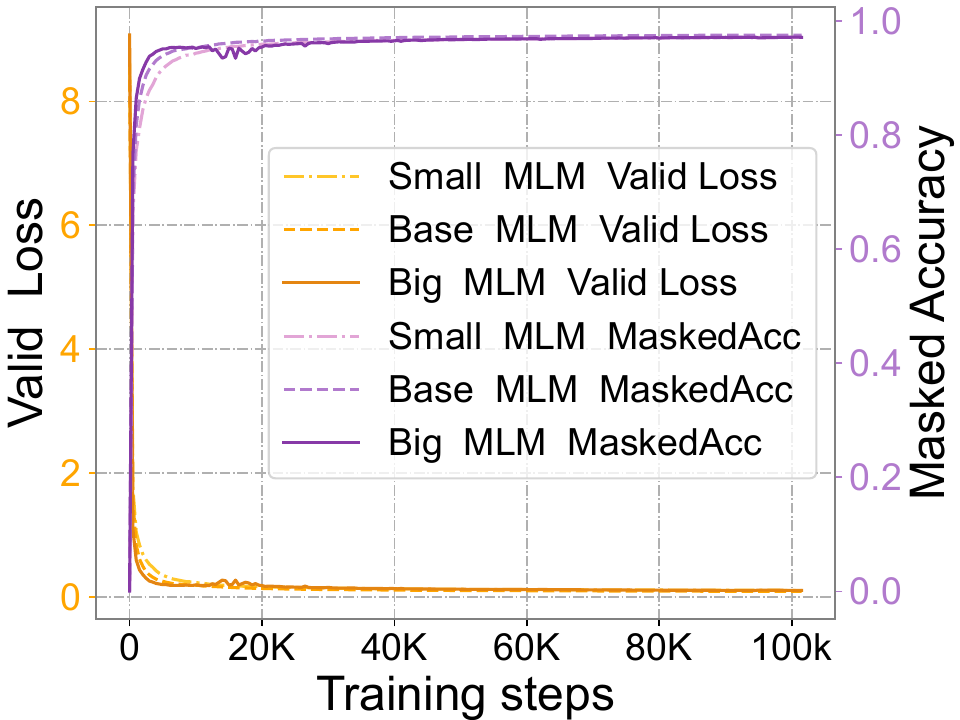}
  }
  \subfloat[Training Curves: Mask Ratios]
  {
      \label{fig:sat_subfig3}\includegraphics[width=0.32\linewidth]{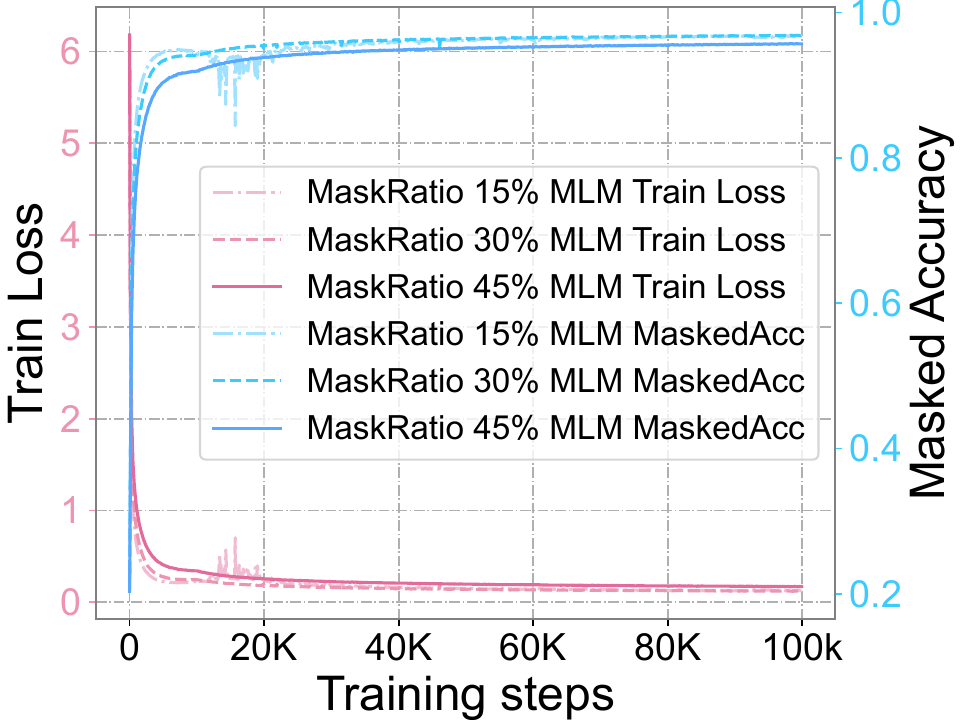}
  }
  \caption{\textbf{Rapid Saturation Problem.} 
We train SMILES MLMs of various sizes and masking ratios using the dataset from \citet{zhou2023uni}. Figure (a) displays the training loss and masking prediction accuracy of different-sized models, showing a rapid decrease in loss and an increase in accuracy at the start of the training. Figure (b) presents similar trends for the validation set. Figure (c) illustrates the training loss and accuracy for models with different masking ratios, showing similar patterns.}    
  \label{fig:saturation}            
  % \vspace{-0.2cm} 
\end{figure*}

\subsection{Rapid Saturation Problem}
\paragraph{Rapid Saturation During Pre-training.} To investigate whether the SMILES MLM model experiences rapid saturation during training and how this issue impacts its scalability, we trained MLMs of various scales and compared their training curves (Details of models with different scale can be found in Appendix \ref{more_pra_model}). As shown in Figure \ref{fig:sat_subfig1}, the training loss decrease rapidly, with the mask-prediction accuracy on the training set exceeding $90\%$ within the first $5{,}000$ steps. By approximately $10{,}000$ steps, all models achieved over $95\%$ accuracy on the training set. A similar rapid saturation phenomenon is observed on the validation set, as shown in Figure \ref{fig:sat_subfig2}, where the validation loss drops sharply after training begins and mask-prediction accuracy rises quickly. The rapid saturation phenomenon is consistent across all models, regardless of scale, including the small model with just  $6.7$M parameters. These findings indicate that the MLM pre-training task is overly simplistic, enabling even very small models to converge quickly, which restricts the models' capacity and scalability for more complex tasks. 
We also test the performance of MLMs of different sizes and different training steps on downstream tasks (detailed results are in Appendix \ref{mlm_down_step}). The downstream-task results further confirm the limited scalability of SMILE MLMs.

\paragraph{Different Mask Ratio Cannot Alleviate Rapid Saturation.} One possible reason for the rapid saturation observed in MLM pre-training is the low masking ratio, with only $15\%$ of tokens masked during training. This might provide insufficient training information and make token prediction too easy, leading to rapid saturation. To explore this hypothesis, we trained large-scale MLMs with different mask ratios (i.e., $15\%$, $30\%$, $45\%$). The training curves, shown in Figure \ref{fig:sat_subfig3}, reveal that MLMs with different mask ratios all exhibit a sharp decline in training loss at the beginning of the training process, converging quickly to a very low level. Even with a mask ratio of $45\%$, the training loss drops rapidly, and by $10$K steps, the mask-prediction accuracy already exceeds $92\%$. These findings indicate that increasing the mask ratio does not mitigate the rapid saturation problem or enhance the scalability of SMILES MLMs. Instead, the results suggest that the rapid saturation stems not from the masking ratio but from the inherent simplicity of the MLM pre-training task, which fails to provide sufficient complexity or information for modeling SMILES data effectively.

\subsection{Challenges in Modeling Substructure Semantics}
\label{analyse_substructure}

To evaluate the ability of MLMs to learn molecular substructure semantics, such as functional groups, we design experiments to analyze whether the models can accurately capture functional group information that is closely related to molecular properties. For this purpose, we use two molecular property prediction datasets, ESOL and FreeSolv \citep{molnet}, both of which are highly relevant to molecular hydrophilicity. Specifically, the ESOL dataset provides information on the water solubility of molecules, while the FreeSolv dataset focuses on the hydrogen free energy, both of which are strongly associated with hydrophilic groups within the molecules.

\begin{figure}[t]    
% \vspace{-1cm} 
  \centering       
  \subfloat[ESOL Dataset]  
  {
      \label{fig:substruct_subfig1_mlm}\includegraphics[width=0.47\textwidth]{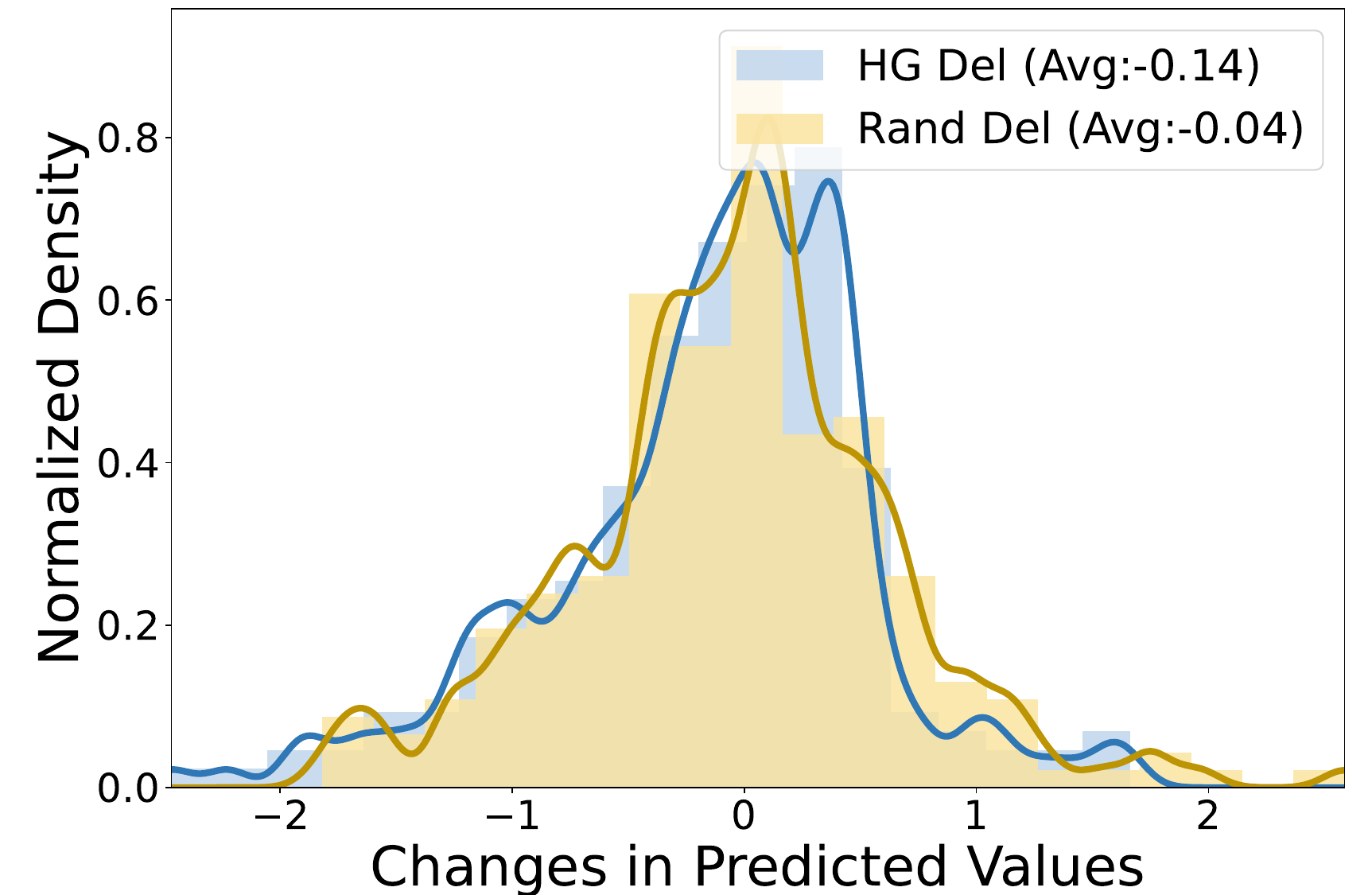}
  }
  \subfloat[FreeSolv Dataset]
  {
      \label{fig:substruct_subfig2_mlm}\includegraphics[width=0.47\textwidth]{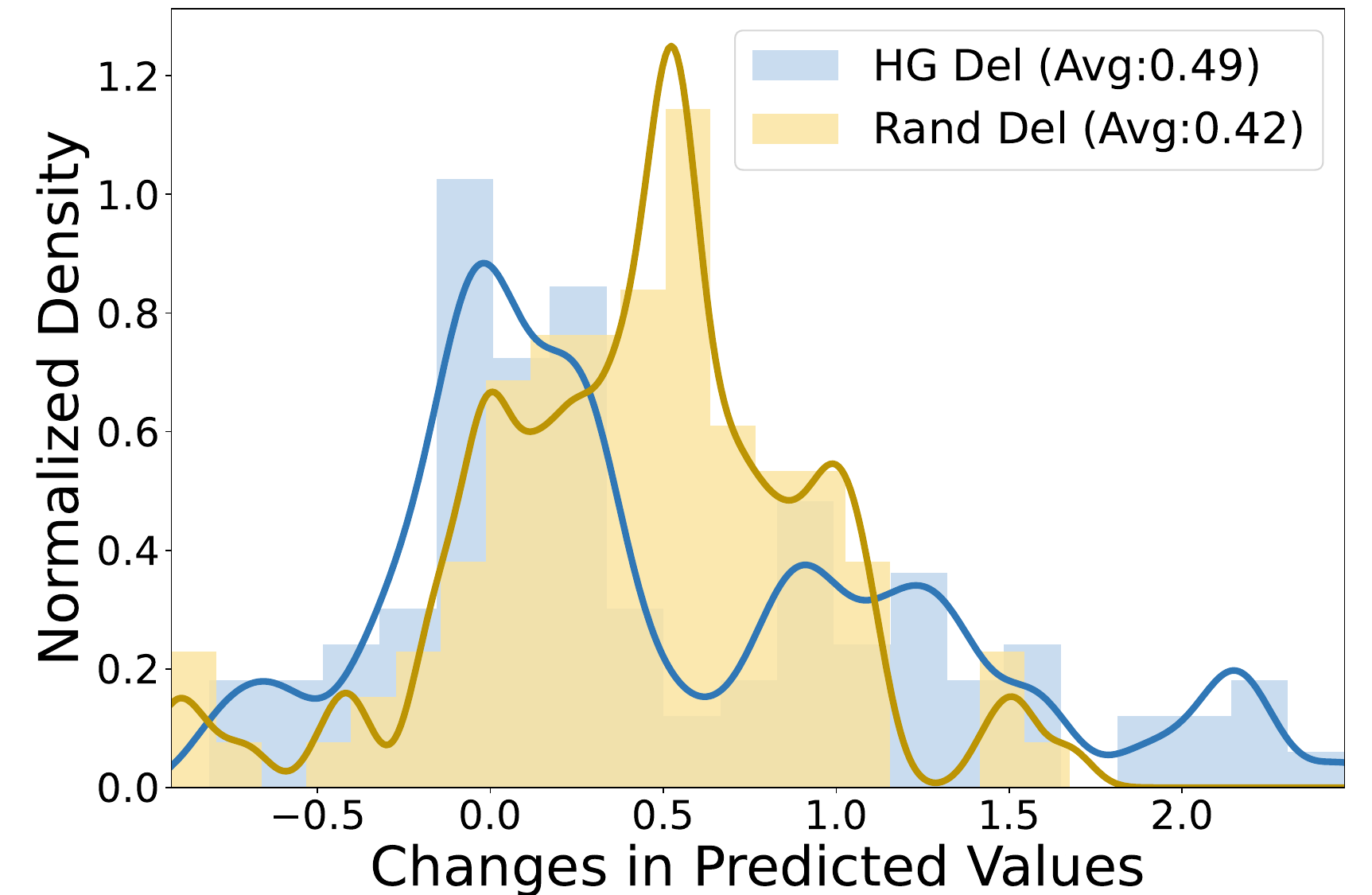}
www.  }

  \caption{\textbf{Substructure Semantics Modeling.} We compared two molecular perturbation methods—removing hydrophilic groups and randomly deleting atoms—and their effects on the model's predictions of hydrophilicity and related properties. Figure (a) presents the impact of these perturbations on model predictions in the ESOL dataset, including the distribution of prediction changes. The average prediction change is similar for both methods (-0.14 vs. -0.04) and shows similar distributions. Figure (b) shows the effects on the FreeSolv dataset, also with similar average prediction change.}    
  \label{fig:substruct}       
  % \vspace{-0.1cm} 
\end{figure}

\begin{figure}[t] 
% \vspace{-0.1cm} 
  \centering        
  \subfloat[ESOL Dataset]  
  {
      \label{fig:substruct_subfig1_levt}\includegraphics[width=0.47\textwidth]{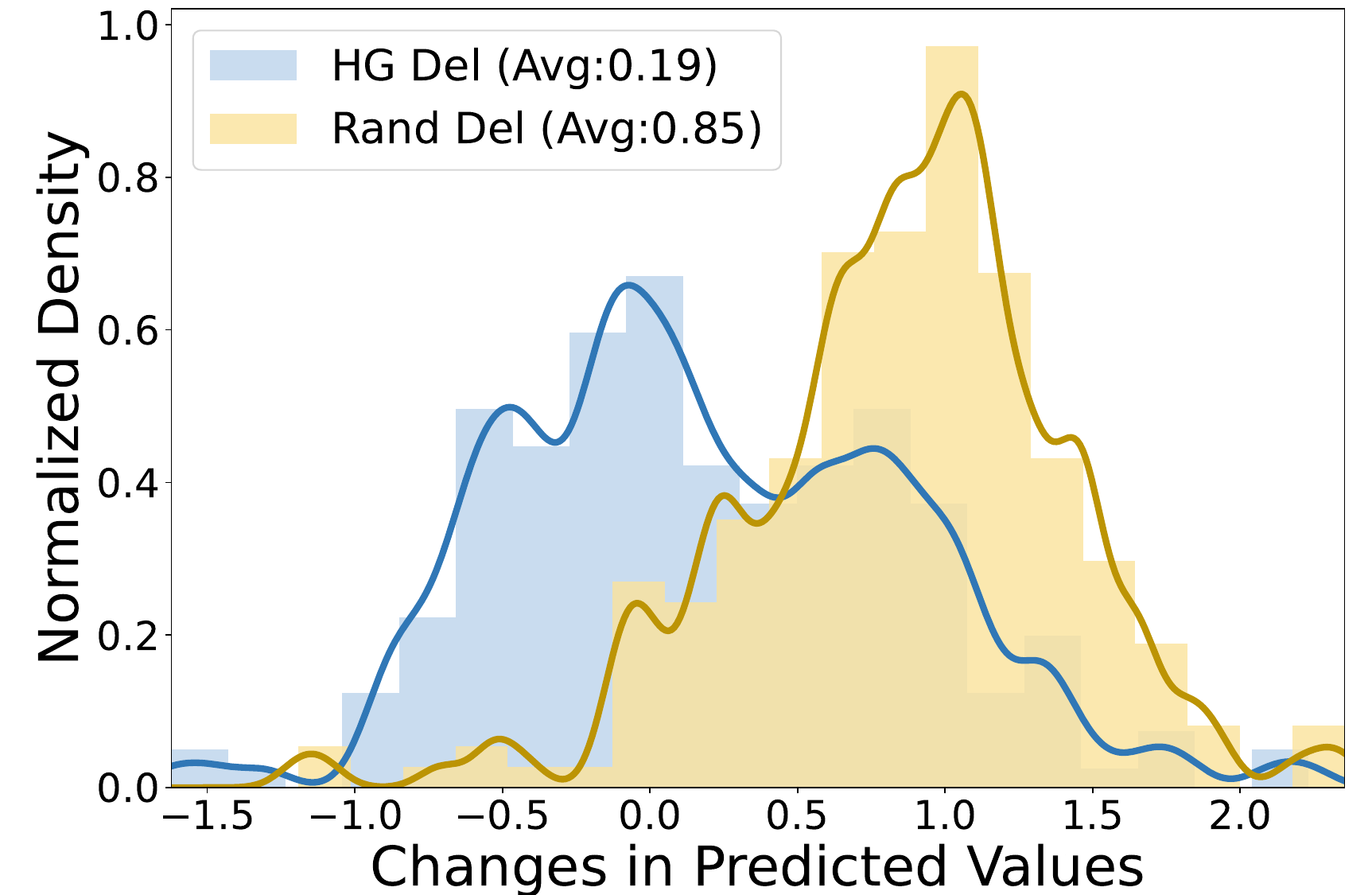}
  }
  \subfloat[FreeSolv Dataset]
  {
      \label{fig:substruct_subfig2_levt}\includegraphics[width=0.47\textwidth]{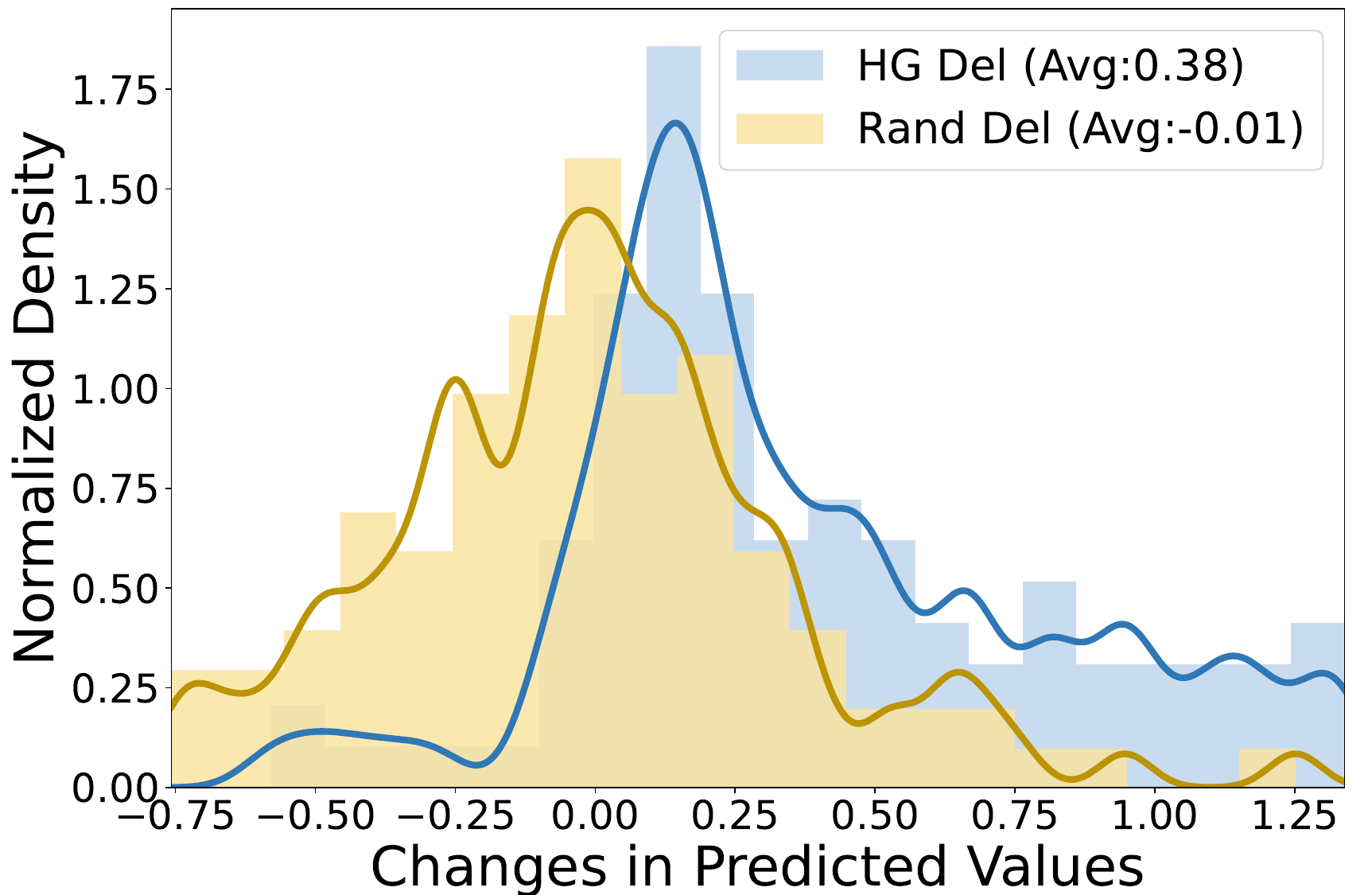}
  }

  \caption{\textbf{Substructure Semantics Modeling from \method{}.} We compared the effects of two molecular perturbation methods on the \method{}'s predictions of hydrophilicity and related properties.  Figure (a) and Figure (b) show that the impact of deleting hydrophilic groups (HG Del) and randomly deleting atoms (Rand Del) on the model's predictions differs significantly, both in the average change in prediction values and their distributions. }    
  \label{fig:substruct_levt}      
% \vspace{-0.5cm} 
\end{figure}

In our approach, we first fine-tune the MLM on these datasets using linear probing. Next, we traverse the SMILES of all molecules in the datasets and remove the hydrophilic groups (e.g., \texttt{-OH}, \texttt{-COOH}, $\texttt{-NH}_{\texttt{2}}$, etc.) identified in each molecule. We then compare the predicted molecular property values before and after the removal. As a control, we also randomly delete atoms from these molecules and analyzed the predicted changes in molecular properties.

As shown in Figure \ref{fig:substruct}, the changes in predicted values after deleting hydrophilic groups (HG Del) are similar to those from random deletions (Rand Del) in both the ESOL and FreeSolv datasets. This indicates that the model struggles to differentiate between the impacts of removing hydrophilic groups and that of random atoms on molecular properties. These results further suggest that the MLM fails to effectively capture the semantic information of key molecular substructures in SMILES.

\section{Edit-based Pre-training Framework}

To overcome the limitations of MLM-based SMILES LMs, we propose a novel SMILES LM that employs an edit-based training objective. To enhance the model's ability to capture substructure semantics, we introduce fragment-level supervision during pre-training. This involves randomly discarding substructures and requiring the model to predict the missing components. This method enables the model to effectively learn substructure semantics. 
MLMs only operate on corrupted SMILES contexts with unreal \texttt{[MASK]} symbols, leading to inconsistencies between training and testing. In contrast, our approach inputs complete and valid SMILES sequences into the model, seeking to reconstruct the missing substructures through an editing-based approach. Moreover, the editing framework offers greater flexibility compared to MLM, as it imposes no specific restrictions on input forms. This flexibility allows us to create more versatile model inputs by removing certain substructures from a molecule, converting it back to SMILES, and then feeding it into the model. In this section, we provide an in-depth discussion of \method{}, focusing on its model design and its pre-training framework.

\subsection{SMILES Encoder with Editing Operations}

\begin{figure*}[t]

% \vspace{-0.9cm}
\centering
	\includegraphics[width=\linewidth]{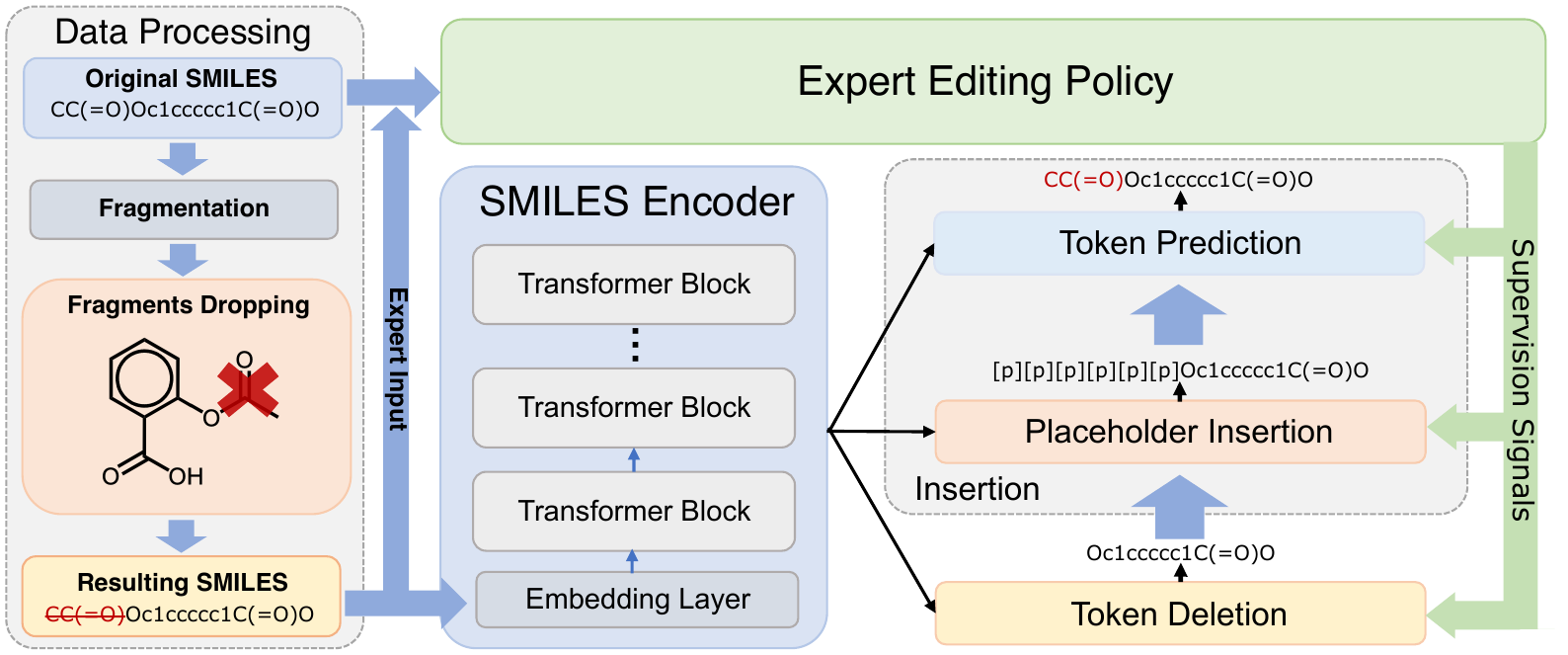}
 
	\caption {\textbf{Overall Framework of \method{}}. The framework includes a data processing module, a SMILES encoder, and an edit-based pre-training process. In data processing, some fragments of the input molecule are randomly removed, and the resulting SMILES is fed into the model. The pre-training goal for the model is to edit the corrupted SMILES to recover the original SMILES. To enable this, we add three different heads for token deletion, placeholder insertion, and token prediction to the SMILES encoder (see Appendix \ref{details_levt} for details). An expert provides training signals for these operations to help the model learn how to recover the original SMILES through editing.}
	\label{fig::framwork}
% \vspace{-0.5cm}
\end{figure*}

In the edit-based pre-training process, the model learn to model the editing operations. Specifically, given a SMILES sequence with missing substructures, the model needs to accurately predict the editing operations required to reconstruct these missing substructures. To achieve this, we designed a SMILES encoder capable of supporting editing-operation modeling.

\paragraph{Model Architecture.} The core architecture of the model is a Transformer Encoder composed of multiple stacked Transformer blocks. Each transformer block contains a multi-head self-attention layer and a feed-forward layer \citep{vaswani2017attention}. The SMILES representations extracted by the Transformer Encoder are then passed to the Editing Operations Head, which predicts the necessary editing operations. Similar to existing Edit-based LMs 
\citep{gu2019levenshtein}, the model handles two types of editing operations: deletion and insertion. Specifically, it completes missing parts of a SMILES sequence by removing redundant parts and inserting missing substructures. To support these tasks, the architecture includes dedicated prediction heads for both editing operations, enabling the model to effectively model editing tasks.

\paragraph{Deletion Operations Head.}  For a given input token, deletion is a binary classification problem -- deciding whether to delete or retain the token. Let $\bm{x}_i^E$ denote the representation of the $i$-th input token extracted by the Encoder. The probability of deleting the $i$-th token, denoted as $\pi^{\rm del}_{\theta}(i)$, can be expressed as:

$$\pi^{\rm del}_{\theta}(i) = \text{Softmax}(\bm{W}_d^T \bm{x}_i^E)\,,$$
where we have the weight matrix $\bm{W}_d$ of size $H \times 2$, and $H$ is the hidden size. 

\paragraph{Insertion Operations Heads.} Modeling insertion operations is more complex than deletion. Inspired by LevT~\citet{gu2019levenshtein}, the insertion operation is split into two steps. In the first step, the model predicts the positions and number of tokens to insert into the original input sequence. Placeholders \texttt{[P]} are then inserted at these positions, representing the locations where new tokens will be added. Later on, in the second step, the model predicts the actual tokens for each placeholder \texttt{[P]}.

Given the ordinary length of SMILES sequences, we constrain the model to insert at most $255$ tokens at a time. Therefore, this step is framed as a $256$-class classification problem for each token position. The probability of inserting tokens at the $i$-th position, denoted as $\pi^{\rm ins}_{\theta}(i)$, is defined as:

$$\pi^{\rm ins}_{\theta}(i) = \text{Softmax}(\bm{W}_{\rm in}^T \bm{x}_i^E)\,,$$
where $\bm{W}_{\rm in}$ is a matrix of size $H \times 256$.

In the second step, for each placeholder symbol \texttt{[P]}, the model predicts the token to insert. This step resembles MLM pre-training tasks, where the model predicts the probability distribution over the vocabulary for the token at each placeholder. The probability distribution for the token at the $i$-th position, denoted as $\pi^{\rm tok}_{\theta}(i)$, is defined as:

$$\pi^{\rm tok}_{\theta}(i) = \text{Softmax}(\bm{W}_{\rm tok}^T \bm{x}_i^E)\,,$$
where $\bm{W}_{\rm tok}$ is a matrix of size $H \times {\rm vob}$, where ${\rm vob}$ represents the size of the vocabulary.

\subsection{Edit-based Pre-training with Fragment-level
Supervision}

After constructing the SMILES encoder with editing operations, the next crucial step is to develop an edit-based pre-training framework that incorporates fragment-level self-supervised training signals. Unlike traditional MLMs, our edit-based model transforms a valid SMILES input into the target SMILES by modeling the necessary editing operations. This process begins by fragmenting the input molecule using rule-based molecule fragmentation, and then randomly selected some fragments to discard from the original molecule. The resulting corrupted molecule is reconstructed into a SMILES representation and fed into the SMILES Encoder.

\paragraph{Molecule Fragmentation and Fragments Dropping.}  To provide the model with fragment-level training signals, we first split the input molecule $M$ into multiple fragments $\{f_1, f_2, ...\}$. A popular method for molecular fragmentation is the BRICS algorithm  \citep{degen2008art}, which divides a molecule into fragments based on predefined rules such as functional groups. However, BRICS often generates relatively large fragments, and removing these fragments can overly disrupt the molecule, potentially eliminating critical structures like rings. To address this, we adopt a modified fragmentation approach on RMCF \citep{wang2022regularized}, which further splits connections between rings and side chains on top of BRICS, resulting in smaller molecular fragments. After cutting the molecule, we randomly select and discard a subset of the fragments with a predefined probability. The remaining fragments are then reassembled into a corrupted molecule \(\hat{M}\). The model is tasked with reconstructing the original molecule's SMILES \(M\) from the given corrupted SMILES \(\hat{M}\) through an edit-based approach.

\paragraph{Edit-based Training Objective with Dual Loss.}  The core pre-training task of the \method{} model is to take the SMILES of a corrupted molecule \(\hat{M}\) as input and attempt to reconstruct the SMILES of the original molecule \(M\) through an editing process, by accurately modeling deletion and insertion operations. However, traditional edit-based models like LevT only provide insufficient training signals for deletion. It only teaches the model to remove incorrect tokens it inserted. This constraint hinders the model's ability to effectively learn how to delete the incorrect parts in the input SMILES. To overcome this problem, we introduce a dual deletion loss, which supervises the model to correctly remove erroneous tokens from the corrupted molecule input \(\hat{M}\).

We adopt the imitation-learning method from LevT, which supervises the model to minimize the Levenshtein distance between the input and target output. The dual deletion objective is defined as:

\[
\mathcal{L}^{\rm DualDel}_{\theta}=-\sum_{y_i \in \hat{M} \atop d_i^* \in d^*} \log \pi_{\theta}^{\rm del}(d_i^*|i,\hat{M})\,,
\]
where \(d^*\) is the optimal deletion action determined by an expert to minimize the Levenshtein distance to the target output \(y^*\) which is the SMILES of molecule $M$. This is formulated as \(d^* = {\rm argmin}_d \mathcal{D}(y^*, \varepsilon(\hat{M}, d))\), where \(\mathcal{D}\) is the Levenshtein distance, \(\pi_{\theta}^{\rm del}\) is the Deletion Classifier, and \(\varepsilon\) represents the environment in LevT’s Markov Decision Process. \(\varepsilon(\hat{M}, d)\) applies the deletion action \(d\) to the initial input SMILES \(\hat{M}\), removing selected tokens. 

In addition to the dual deletion loss, we retain LevT's original training objective \(\mathcal{L}^{\rm LevT}_{\theta}\) (see Appendix \ref{details_levt} for details). This supervises both deletion and insertion actions by minimizing the Levenshtein distance between the input \(\hat{M}\) and target \(M\). The final training objective for \method{} is a combination of these two losses:

\[
\mathcal{L}_{\theta} = \mathcal{L}^{\rm DualDel}_{\theta} + \mathcal{L}^{\rm LevT}_{\theta}\,.
\]

Through this edit-based pre-training process, we equip the \method{} model with fragment-level training signals, enabling it to reconstruct the SMILES of a corrupted molecule \(\hat{M}\) into the SMILES of the original molecule \(M\) via fragment editing.

\section{Experiments}

In this section, we first evaluate the performance of \method{} on molecular property prediction tasks and compare it with baseline models (see Section \ref{sec:main_res}). The results show that \method{} outperforms both the MLM and 3D molecular models, achieving state-of-the-art performance. To further validate the model design and pre-training framework, we conduct ablation studies on training signals and editing operations (see Section \ref{sec:ablation}). In addition, analytical experiments confirm that \method{} has a stronger ability to capture the semantics of molecular substructures compared to MLMs. Analysis of the training curves also demonstrates that \method{} mitigates the issue of rapid saturation and enhances the training stability (Section \ref{sec:analuse_res}). We also provide details on the training data, baseline models, and implementation used in the experiments (see Section \ref{sec:settings}).

\subsection{Experiment Settings}
\label{sec:settings}
\paragraph{Datasets.} For pre-training, we use the large-scale molecular dataset provided by \citet{zhou2023uni}, which includes SMILES information for $19$ million molecules. For fine-tuning, we employ the widely-recognized MoleculeNet benchmark \citep{wu2018moleculenet} (see Appendix \ref{sec::ft_datasets} for more details). We follow the same data split as used by \citet{zhou2023uni} and tokenize SMILES sequences with the regular expression from \citet{schwaller2018found}.

\paragraph{Baselines.} We evaluate our approach against various supervised learning and pre-training baselines, including both SMILES-based and 3D molecular pre-training models. The supervised methods include D-MPNN \citep{yang2019analyzing} and AttentiveFP \citep{xiong2019pushing}, both of which are based on graph neural networks (GNNs). For 2D and 3D molecular pre-training, we consider baseline methods: N-gram \citep{liu2019n}, GROVER \citep{rong2020self}, GraphMVP \citep{liu2021pre}, MolCLR \citep{wang2022molecular}, InfoGraph \citep{sun2019infograph}, Mole-BERT \citep{xiamole}, 3D InfoMax \citep{stark20223d}, and MoleculeSDE \citep{liu2023group}. For a fair comparison, we train a SMILES model based on MLM pre-training, referred to as {\mlmmth}, using the same training data, model architecture, and training hyperparameters as \method{}.

\paragraph{Implementation Details.} We use a Transformer block with a hidden size of $768$ and $12$ attention heads, comprising $12$ layers in the SMILES encoder, which contains a total of $86.3$ million trainable parameters. During pre-training, the fragment drop ratio is set to $0.15$. For downstream tasks, we use the same fine-tuning dataset established by Uni-Mol.  (cf. Appendix \ref{sec::pretrain_configuration} for more details about hyper-parameter configuration.)

\subsection{Results on Molecular Property Classification Tasks}
\label{sec:main_res}

\begin{table*}[t]
% \vspace{-0.8cm}
\centering
\caption{The overall results on $7$ molecule property classification datasets. We report ROC-AUC
score (higher is better) under scaffold splitting. The best results are \textbf{bold}. The second-best results are \underline{underlined}. For more detailed information about the dataset, please refer to Table \ref{fine_tune_dataset}.
}
\footnotesize
\begin{tabular}{c|ccccccc|c}

\toprule
 Datasets    &   BACE$\uparrow$&   BBBP$\uparrow$&   Tox21$\uparrow$&  SIDER$\uparrow$&     MUV$\uparrow$&    ClinTox$\uparrow$&    ToxCast$\uparrow$ & Mean$\uparrow$\\
 \# Molecules &  1531&   2039&   7831&  1427&      93087&     1478&    8575 &-\\
 \midrule
 D-MPNN &   80.9&   71.0&   75.9&  57.0&      78.6&       90.6&    65.5 &  74.2\\
 Attentive FP &   78.4&   64.3&   76.1&  60.6&     76.6&      84.7&    63.7 & 72.1\\
 N-Gram$_{\mathrm{RF}}$ & 77.9&   69.7&   74.3&  \textbf{66.8}&    76.9&      77.5&  - & - \\
 GROVER&   \textbf{82.6}&   70.0&   74.3 &  \underline{64.8}&  62.5&          81.2&    65.4 & 71.5\\
 GraphMVP &   81.2&   \underline{72.4}&   75.9&  63.9&     77.7&       79.1&    63.1 & 73.3\\
InfoGraph&   77.8&   67.5&   73.2&  59.9&     74.1&       76.5&    63.7 & 70.4 \\
 MolCLR &   \underline{82.4}&   72.2&   75.0&  58.9&     79.4&       \underline{91.2}&    \textbf{69.2} & \underline{75.5} \\
Mole-BERT & 80.8&   71.9&   \underline{76.8}&  62.8&     78.6&       78.9&    64.3 & 73.4
 \\
 3D InfoMax & 79.7&   69.1&   74.5&  60.6&      74.4&       79.9&    64.4 & 71.8 \\
  MoleculeSDE & 80.4&   73.2  & 76.5 &  59.6&      \underline{79.9}&       86.6&    65.2 & 74.5 \\
 \midrule 
\mlmmth & 77.8&   68.6&   75.1&  61.2&      75.1&      89.8&    64.9 & 73.2 \\
 \rowcolor{mygray} \method{} & 80.3 & \textbf{77.4} & \textbf{77.1} &   63.0& \textbf{80.2} & \textbf{98.9} & \underline{67.4} & \textbf{77.8}\\
\bottomrule
\end{tabular}
\label{table::classification}
% \vspace{-0.5cm}
\end{table*}

\paragraph{Tasks Details.} We evaluate \method{} on the MoleculeNet \citep{molnet} benchmark and compare its performance with baseline models. We evaluate \method{} on $7$ widely-used molecular property prediction tasks ( see Appendix \ref{sec::ft_configuration} for details). For all the seven tasks, we take the normalized SMILES information as model input and fine-tuning on each task separately. We use ROC-AUC as the evaluation metric, and the results are summarized in Table \ref{table::classification}. 

\paragraph{Results.}  \method{} achieves state-of-the-art (SOTA) performance on $4$ out of $7$ tasks, and closely matches the SOTA models on the remaining tasks. Compared to the MLM model {\mlmmth}, whose training settings, training data, and evaluation processes for downstream tasks are all identical to ours, \method{} demonstrates superior performance across all seven tasks. These results validate the effectiveness of our pre-training framework. Additionally, \method{} outperforms several molecular representation learning models that utilize 3D information, indicating that SMILES has the potential of revealing important and rich semantic information related to the spatial molecular properties, and that \method{} effectively captures such information. \method{} also demonstrated the strongest average performance across all $7$ tasks, indicating that it outperforms other baseline models in these prediction tasks overall.

\subsection{Ablation Studies}
\label{sec:ablation}

\subsubsection{Ablation Studies on Fragment-level Supervision}

\begin{table*}[ht]
% \vspace{-0.2cm}
\centering
\caption{\textbf{Ablation Studies on Fragment-level Supervision.} 
Fragment-level supervision provide more informative and useful training signals than atom-level supervision and are crucial for helping the model learn multi-level semantic information in molecules.
}
\footnotesize
\begin{tabular}{c|ccccc|c}
\toprule
 Method    &   BACE$\uparrow$&   BBBP$\uparrow$&   Tox21$\uparrow$&  SIDER$\uparrow$&    ToxCast$\uparrow$ & Mean$\uparrow$\\
 \midrule
\method{}-AtomsDropping & 80.0 & \underline{73.4} &  \underline{76.5} &  59.2& \underline{66.6} & \underline{71.1}\\
\method{}-AtomsMasking & \textbf{80.4} & 73.2 &  75.0 &  58.3& 64.6 & 70.3\\
 \mlmmth & 77.8 & 68.6 &  75.1 &  \underline{61.2}& 64.9 & 69.5 \\
 \midrule    
 \rowcolor{mygray} \method{} & \underline{80.3} & \textbf{77.4} & \textbf{77.1} &   \textbf{63.0}& \textbf{67.4} & \textbf{73.0}\\

\bottomrule
\end{tabular}
\label{table::ablation_input}
% \vspace{-0.5cm}
\end{table*}

\paragraph{Experimental Settings.} To explore the impact of fragment-level supervision signals versus the atom-level signals on model performance, we additionally train \method{}'s variants using two different pre-training strategies. The first model, \method{}-AtomsDropping, replaces the fragment dropping process in pre-training with random atom dropping. After discarding certain atoms, we input the modified SMILES into the model, asking it to restore the original SMILES through an editing approach. The second model, \method{}-AtomsMasking, uses random token masking similar to MLM, where selected tokens are replaced with \texttt{[MASK]}, and the model is tasked with restoring the original SMILES via editing. The performance of these models is presented in Table \ref{table::ablation_input}.

\paragraph{Results Analysis.} The results show a significant decline in performance when fragment-dropping is replaced with random-atom-dropping (\method{}-AtomsDropping vs. \method{}), indicating that the fragment-level supervision signal enables the model to learn more important and nuanced semantic information. Furthermore, when random atom dropping is replaced with random token masking, performance decreases again (\method{}-AtomsMasking vs. \method{}-AtomsDropping). This suggests that while both random-atom-masking and random-atom-dropping introduce atom-level training signals, the introduction of the unrealistic special symbol \texttt{[MASA]} through token masking adversely affects model performance. Compared to these two models, {\mlmmth} exhibits even poorer performance, demonstrating that this editing training approach effectively helps the model learn richer semantic knowledge.

\subsubsection{Ablation Studies on Editing Operations}

\begin{table*}[ht]
% \vspace{0.5cm}
\centering
\caption{\textbf{Ablation Studies on Editing Operations.} The placeholder insertion process, which is absent in MLMs, enables the model to learn richer and more diverse semantic information.
}
\footnotesize
\begin{tabular}{c|ccccc|c}
\toprule
 Method    &   BACE$\uparrow$&   BBBP$\uparrow$&   Tox21$\uparrow$&  SIDER$\uparrow$&    ToxCast$\uparrow$ & Mean$\uparrow$\\
 \midrule
w/o PlhIns & 76.1 & 69.7 & 76.9 &  55.5& \underline{66.2} & 68.9\\
w/o TokPred & \underline{79.8} & 69.2 &  75.4 &  57.4& 65.9 & 69.5\\
w/o TokDel & 79.0 & \underline{73.5} &  \textbf{77.3} &  \underline{61.9}& 64.9 & \underline{71.3}\\
w/o DualDel & 78.4 & 70.1 &  76.4 &  59.5& 64.4 & 69.8\\
 \midrule    
 \rowcolor{mygray} \method{} & \textbf{80.3} & \textbf{77.4} & \underline{77.1} &   \textbf{63.0}& \textbf{67.4} & \textbf{73.0}\\
\bottomrule
\end{tabular}
\label{table::ablation_edit}
% \vspace{-0.5cm}
\end{table*}

\paragraph{Experimental Settings.} To investigate the impact of different training signals from the editing operations in the \method{} model on its performance, we train four additional variants of the \method{} model. These models are obtained by removing the training signals for placeholder insertion, token prediction, token deletion, and dual token deletion (setting the training loss weight to zero), corresponding to the three editing operations in the original LevT model and the dual deletion loss. The detailed results are presented in Table \ref{table::ablation_edit}.

\paragraph{Results Analysis: Why \method{} is Better than {\mlmmth}.}  The results indicate that the ablation of any of these four editing operations leads to a significant decline in model performance. Notably, removing the placeholder insertion operation results in the largest performance drop. This operation primarily models the position of missing fragments within the SMILES, highlighting the importance of teaching the model to predict the locations of these fragments for capturing critical semantic information and improving performance. In contrast, the MLM model attempts to predict masked tokens based on their given positions, which simplifies the pre-training task and limits the model’s exposure to important semantic information, ultimately affecting its performance. Moreover, \method{} provides supervision signals for each token in the sequence, but the MLM model only provides supervision signals for \texttt{[MASK]} tokens, limiting the semantic richness of the model.

\paragraph{Results Analysis: Dual Deletion Loss is More Useful.} The ablation of the dual deletion operation also causes a significant decline in model performance, with a more pronounced drop than when token deletion is removed. This indicates that the dual-deletion-objective of {\method} provides more useful and richer training signals than token-deletion objective in LevT.

\subsection{Analytical Experiments}
\label{sec:analuse_res}

\paragraph{\method{} Understands Substructure Semantics.} Similar to the analysis in Section \ref{analyse_substructure}, we tested \method{}'s response to two different molecular perturbation methods on the ESOL and FreeSolv datasets. As shown in Figure \ref{fig:substruct_levt}, compared it to the results in Figure \ref{fig:substruct}, \method{} exhibits distinct prediction changes for the two perturbation methods on both the ESOL and FreeSolv datasets. This indicates that \method{} can clearly differentiate between the impact of removing hydrophilic groups and randomly-selected atoms on molecular properties, demonstrating that it models the semantics of key molecular substructures more effectively than the MLM.

% \begin{figure}
%     \centering
%     \includegraphics[width=0.5\linewidth]{pics/loss_100K_train_levt.pdf}
%     \caption{The training loss curves of different-sized \method{} models. The loss curves consistently show a stable downward trend throughout the training process, and the model loss gradually decreases as the model size increases.}
%     \label{fig:train_levt}
% \end{figure}

\begin{figure}[t] 
% \vspace{-0.1cm} 
  \centering        
  \subfloat[Training Curves]  
  {
      \label{fig:substruct_subfig1_levt_scale}\includegraphics[width=0.47\textwidth]{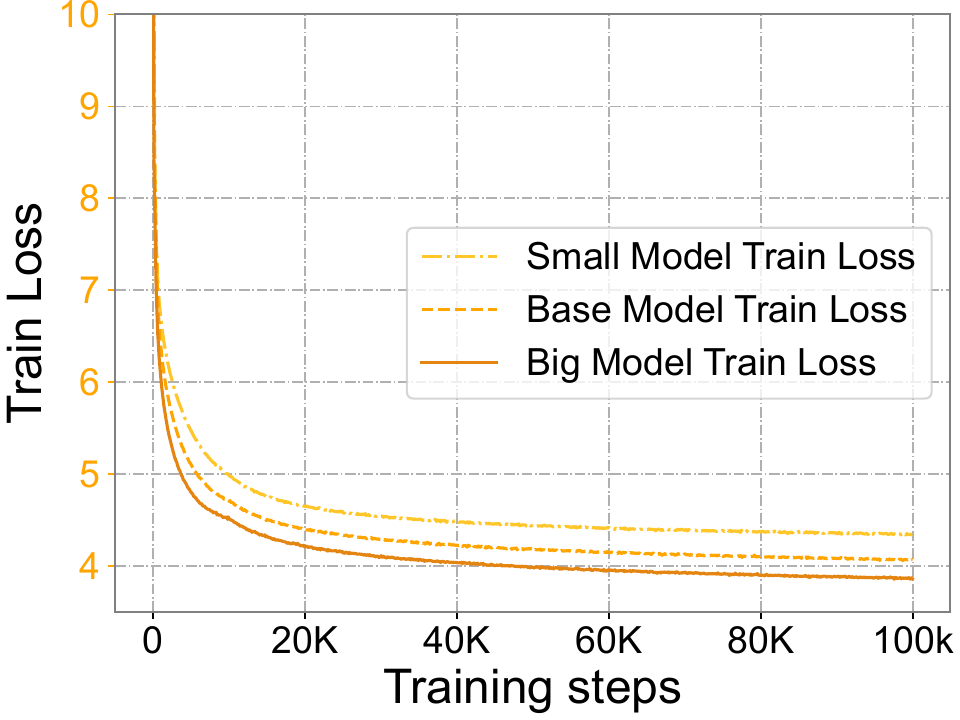}
  }
  \subfloat[Validation Curves]
  {
      \label{fig:substruct_subfig2_levt_scale}\includegraphics[width=0.47\textwidth]{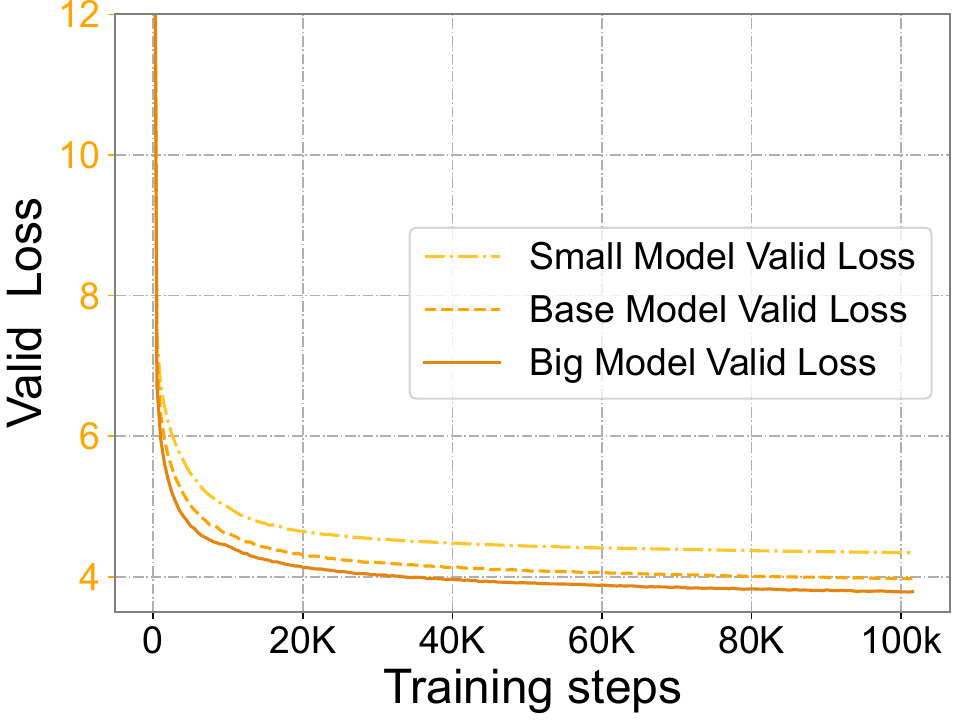}
  }

  \caption{The training and validation loss curves of different-sized \method{} models. The loss curves consistently show a stable downward trend throughout the training process, and the model loss gradually decreases as the model size increases.}    
  \label{fig:train_levt}      
\end{figure}

% \begin{wrapfigure}{r}{0.4\textwidth}
% % \vspace{-2.5em}
% \begin{center}
% \includegraphics[width=\linewidth]{pics/loss_100K_train_levt.pdf}
% \end{center}
% \vspace{-1.5em}
% \caption{The training loss curves of different-sized \method{} models. The loss curves consistently show a stable downward trend throughout the training process, and the model loss gradually decreases as the model size increases.}
% \label{fig:train_levt}
% \vspace{-1.5em}
% \end{wrapfigure}

\paragraph{\method{} Enhances Training Stability and Model Scalability.} We train \method{} of different sizes and compare their training curve variations. As shown in Figure \ref{fig:train_levt}, the losses of the \method{} models consistently exhibit a more pronounced downward trend throughout the training process compared to the MLMs (see Figure \ref{fig:sat_subfig1}), further  alleviating the rapid saturation problem. Additionally, unlike the MLM, the training loss of the \method{} shows more distinct differences across sizes. As the model size increases, the loss steadily decreases, with the larger model (Big Model) converging more stably than the MLM, indicating better scalability for \method{}. We also analyze the training and validation loss curves for the three types of editing operations in \method{}, confirming the model's scalability during pre-training (see Appendix \ref{more_levt_curve} for details). Additionally, we evaluate the performance of \method{} models of different sizes on downstream tasks, demonstrating that \method{} exhibits better scalability and stability compared to the MLM (i.e., \mlmmth) (see Appendix \ref{levt_down_step} for details).

\section{Conclusions}

In this paper, we analyze the behavior and shortcomings of masked language models (MLMs) on SMILES data. By examining the training curves, we demonstrate that training MLMs on SMILES data encounters rapid saturation issues. Further analytical experiments reveal that MLMs struggle to effectively capture the semantics of important molecular substructures. To address these issues, we propose the edit-based pre-training molecular representation learning framework \method{}, which is specifically designed to capture substructure semantics by learning how to recover the missing fragments through edit operations. Extensive experiments on molecular property prediction tasks validate the effectiveness of \method{}, and ablation studies confirm the advantages of its design over traditional MLMs in modeling molecular substructure semantics and training stability.

\section{Acknowledgements}
We would like to thank Kuangqi Zhou from Bytedance for his insightful discussions on the project.
We also thank other members from Dlib in Peking University for their valuable feedback given during the internal discussions.
This paper is partially supported by National Natural Science Foundation of China with Grant Number 62276002.

\bibliography{iclr2025_conference}
\bibliographystyle{iclr2025_conference}

\clearpage

\appendix

\section{Related Works}
\label{related_works}
Simplified Molecular Input Line Entry System (SMILES) has emerged as a cornerstone sequential representation for molecular data, making it a focal point in molecular representation learning. A large range of Pre-trained SMILES LMs \citep{wang2019smiles,chithrananda2020chemberta,ross2022large} have been developed to address challenges in SMILES-based molecular modeling, demonstrating their effectiveness across diverse downstream tasks \citep{bagal2021molgpt,tong2021generative,feng2024bioactivity, yang2024poisoning}. 
These models typically leverage techniques like MLM and autoregressive pre-training to capture the complex syntax and semantics embedded in SMILES sequences.
In parallel, edit-based generative models, another important approach to sequence modeling, have gained prominence in broader sequence modeling tasks, including machine translation, text summarization, and grammatical error correction. 
This section provides a comprehensive overview of these two prominent approaches.
We first introduce representative work in Pre-trained SMILES LMs, highlighting their architectures and applications. Then, we discuss edit-based LMs, detailing their methodologies and their relevance to sequence modeling tasks.

\subsection{Pre-trained SMILES Language Model}

Similar to text, SMILES sequences encode information in a sequential format. Early pre-trained SMILES LMs adopted methods techniques originally designed for text modeling. \citet{wang2019smiles} introduced SMILES-BERT, inspired by the BERT model \citep{devlin2018bert}, employing its MLM objective, which has been widely applied in various data representation learning tasks \citep{lin2023evolutionary, zhengesm, yang2024mol, hayes2025simulating}. Their results demonstrated significant improvements in molecular property prediction tasks. Likewise, \citet{chithrananda2020chemberta} developed ChemBERTa, leveraging the advanced RoBERTa architecture \citep{liu2019roberta} to enhance SMILES semantics modeling through MLM pre-training. Further advancing this line of research, \citet{ross2022large} proposed Molformer, trained on a larger dataset with MLM training objective, showing that SMILES LMs can effectively capture both molecular properties and structural patterns. These contributions have established MLM-based methods as a dominant approach in SMILES representation learning. In parallel, generative pre-training approaches have also gained traction in this domain. MolGPT \citep{bagal2021molgpt} employs an autoregressive mechanism, while \citet{tong2021generative} applied generative models to drug design tasks. More recently, \citet{liu2023molxpt} introduced MolXPT, unifying SMILES and textual data using a generative pre-training strategy. Overall, these works highlight the growing importance of pre-trained SMILES LMs, particularly those based on the MLM objective.

\subsection{Edit-based Language Model}
Edit-based sequence generation offers a faster and more flexible alternative to traditional autoregressive methods. \citet{malmi2019encode} introduced the {LASERTAGGER} model, which uses tags (i.e., keep, delete, add) to edit sequences. The Felix model \citep{mallinson2020felix} extended this idea by integrating a pointer-based mechanism with an MLM backbone to handle insertion and deletion tasks more efficiently. 
Recognizing that edit operations from an input sequence to a target output can be diverse and difficult to compute directly, \citet{gu2019levenshtein} developed the Levenshtein Transformer (LevT) model, which calculates the minimum levenshtein distance between the input and target sequences. By leveraging this metric, LevT generates an optimal sequence of edit operations, improving performance on tasks such as machine translation and post-editing. LevT was further applied to lexically constrained translation tasks with notable success \citep{susanto2020lexically}. To resolve inconsistencies between training and inference, \citet{zheng2023towards} introduced a dual training objective, improving the performance of edit-based models in applications such as summarization and grammatical error correction. With their efficiency and adaptability, edit-based LMs have emerged as a highly promising paradigm for sequence modeling across diverse tasks.

\section{Details of Levenshtein Transformer}
\label{details_levt}

The Levenshtein Transformer (LevT) is a non-autoregressive edit-based generation model that employs a three-step editing process: token deletion, placeholder insertion, and token prediction. LevT is trained using imitation-learning with a dual policy: (i) learning to insert tokens by predicting those that have been randomly deleted from the complete sequences, and (ii) learning to delete tokens by identifying incorrect tokens generated by an insertion module. Below are more details about the training objective of LevT, denoted as \(\mathcal{L}^{\rm LevT}_{\theta}\).

\noindent{\bf Placeholder Insertion Loss.} In this step, the model needs to determine how many placeholders \texttt{[P]} should be inserted at specific positions in the original input, which will later be replaced by concrete tokens in subsequent steps. This operation is modeled as a classification task that predicts how many words need to be inserted after each token in the input sequence. For practical implementation, LevT limits the maximum number of words that can be inserted after each token to $255$. Thus, this step essentially becomes a $256$-class classification task at each token, predicting the number of words ($0$--$255$) to insert after each token. This process can be represented as follows:

\begin{eqnarray}
\mathcal{L}^{\rm ins}_{\theta}=-\sum_{y_i\in Y_{0}\atop p_i^*\in p^* } \log \pi_{\theta}^{\rm ins}(p_i^*|i,Y_{0}) \nonumber \,,
\end{eqnarray}
where $p_i^*$ is the optimal placeholder insertion action found by the expert that minimizes the Levenshtein distance to the target output $y^*$ which is the SMILES of molecule $M$, and can be formalized as $p_i^* ={\rm argmin}_p \mathcal{D}(Y^*, \varepsilon(Y_{0}, p))$, $Y_0$ is the initial input of the model which is the SMILES of molecule $\hat{M}$ , $\mathcal{D}$ is the Levenshtein distance measurement, $\pi_{\theta}^{\rm del}$ is LevT's Deletion Classifier, and $\varepsilon$ is the environment in the Markov Decision Process of LevT which receives editing actions and returns the modified sequence, and $\varepsilon(Y_{0}, p)$ means applies the insertion action $p$ to the initial input sequence $Y_{0}$ (e.g., insert some placeholders in $Y_{0}$). Details of $\varepsilon$ can be found in LevT's framework ~\citep{gu2019levenshtein}.

\noindent{\bf Token Prediction Loss.} In this step, the task is to predict an exact token for each placeholder \texttt{[P]} in the sequence \(Y_1=\varepsilon(Y_{0}, p^*)\) that has had placeholders inserted. This process closely resembles that of MLM, as it essentially entails a classification problem where the number of classes is equal to the size of the vocabulary.
\begin{eqnarray}
\mathcal{L}^{\rm tok}_{\theta}=-\sum_{y_i\in Y_1, t_i^*\in t^*\atop y_i=<{\rm [P]}> } \log \pi_{\theta}^{\rm tok}(t_i^*|i,Y') \nonumber\,,
\end{eqnarray}
where $t_i^*$ is the optimal insertion action found by the expert that minimizes the Levenshtein distance to the target output $Y^*$, $Y_1$ is the modified sequence by applying the optimal placeholder action $p^*$ to the input sequence $Y_0$, and these terms can be formalized as: $t_i^*={\rm argmin}_t \mathcal{D}(y^*, \varepsilon(Y_1,t))$, $y_1=\varepsilon(Y_{0}, p^*)$, $d^*$ or $p^* = {\rm argmin}_{d,p} \mathcal{D}(Y^*, \varepsilon(Y_{0}, \{d,p\}))$.  $\pi_{\theta}^{\rm tok}$ is token classifier.

\noindent{\bf Token Deletion Loss.} In the previous insertion step, the model may have inserted incorrect tokens. So in this step, it needs to predict which of the previously-inserted tokens are incorrect and should be deleted. Essentially, this step involves learning how to ``correct'' the errors made during the insertion phase. Specifically, the input to this step is the output from the insertion module, \(Y_2=\varepsilon(Y_1, t)\), where \(t\) represents the actions predicted by the model in the token prediction step. Since the task is to decide whether each token in \(Y_2\) should be deleted, this step is essentially a binary classification task on each token, represented as:
\begin{eqnarray}
\mathcal{L}^{\rm del}_{\theta}=-\sum_{y_i\in Y_{2}\atop d_i^*\in d^* } \log \pi_{\theta}^{\rm del}(d_i^*|i,Y_{2}) \nonumber\,,
\end{eqnarray}
where $d_i^*$ is the optimal delete action found by the expert that minimizes the Levenshtein distance to the target output $Y^*$ which is the SMILES of molecule $M$, and can be formalized as $d_i^* ={\rm argmin}_d \mathcal{D}(y^*, \varepsilon(Y_{2}, d))$ , $\pi_{\theta}^{\rm del}$ is LevT's deletion classifier.

\noindent{\bf Total Loss.} Since the editing process of LevT consists of three steps—token deletion, placeholder insertion, and token prediction—the overall training objective of LevT is the sum of the training objectives for these three processes:
\begin{eqnarray}
\mathcal{L}^{\rm LevT}_{\theta}=\mathcal{L}^{\rm ins}_{\theta} + \mathcal{L}^{\rm tok}_{\theta} + \mathcal{L}^{\rm del}_{\theta} \nonumber\,.
\end{eqnarray}

In summary, unlike MLMs that provide training signals only for each \texttt{[MASK]} symbol in the input sequence, the LevT model offers training signals for every token in both the Placeholder Insertion and Token Deletion steps. This design requires the model to determine whether each token in the input sequence should be deleted and whether new tokens should be inserted after each existing token, thus providing richer semantic information to the model.

\section{More training curves of \method{}}
\label{more_levt_curve}

\begin{figure}[ht]    
% \vspace{-0.5cm} 
  \centering         
  \subfloat[Token Deletion]  
  {
      \label{fig:mtc_tokdel_train}\includegraphics[width=0.32\textwidth]{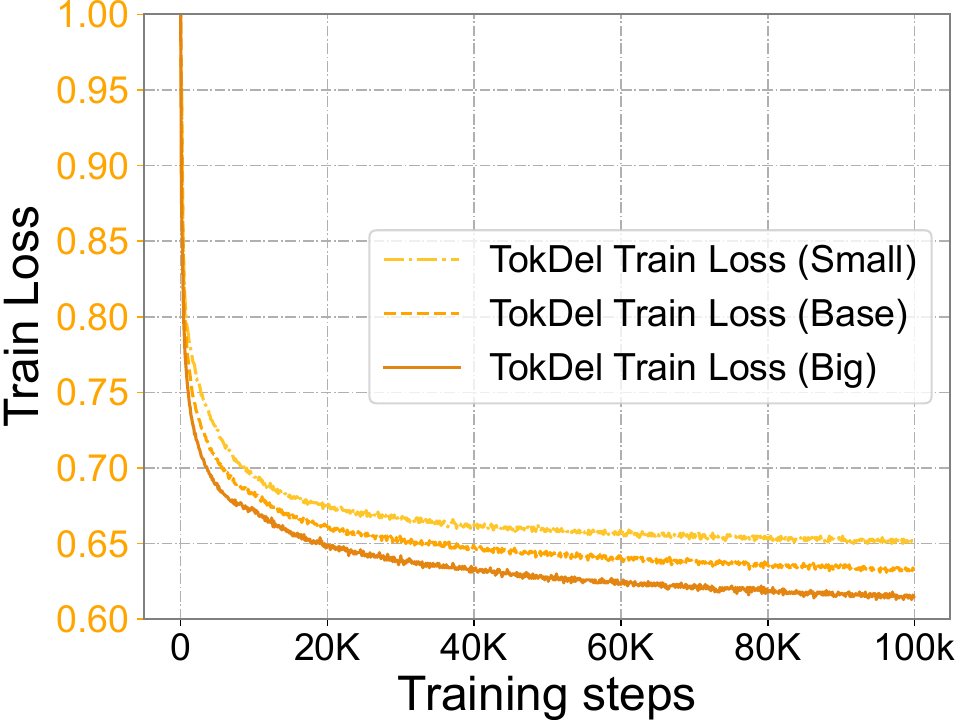}
  }
  \subfloat[Placeholder Insertion]
  {
      \label{fig:mtc_plhins_train}\includegraphics[width=0.32\textwidth]{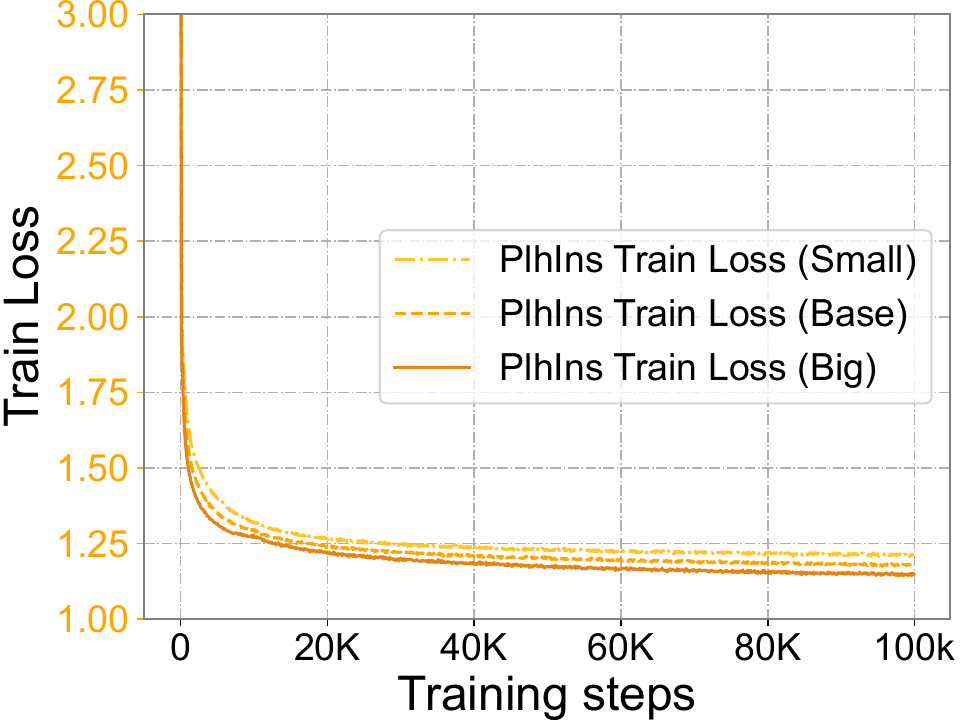}
  }
  \subfloat[Token Prediction]
  {
      \label{fig:mtc_tokpred_train}\includegraphics[width=0.32\textwidth]{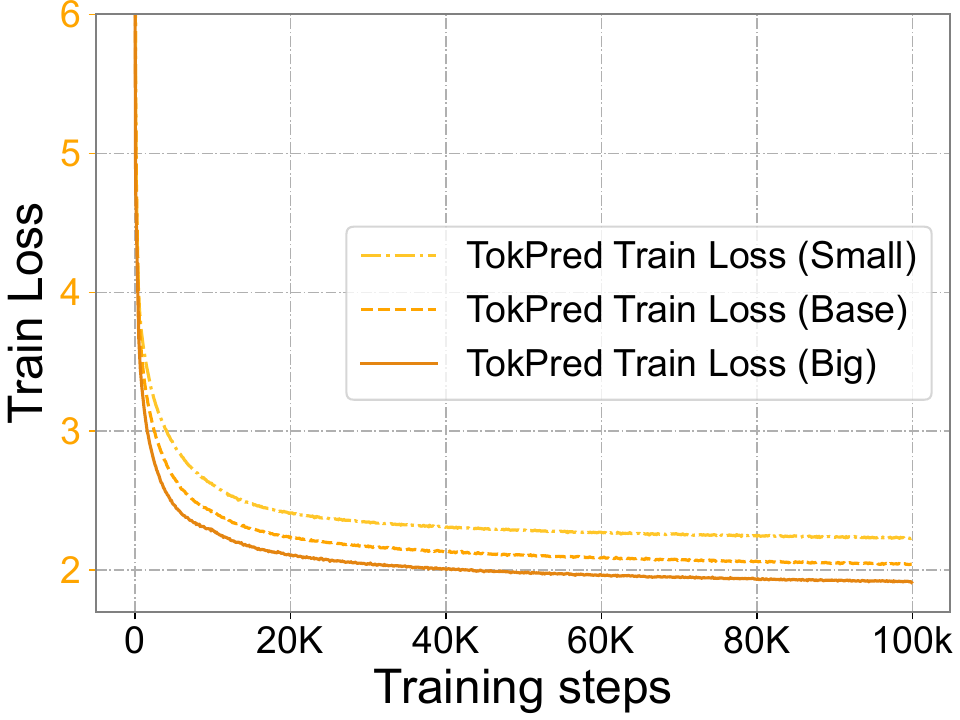}
  }

  \caption{\textbf{Training Loss Curves of Editing Operations.} We train \method{} models of varying sizes and compare their loss curves during training for three different editing operations. As shown in the results, the loss for the token prediction process represented in Figure (c) is consistently the highest among the three type of losses, while the loss for token deletion is the lowest. Furthermore, as the model size increases, all three types of loss exhibit a stable downward trend.}    
  \label{fig:more_train_curve_levt}            
\end{figure}

\begin{figure}[htbp]    
% \vspace{-0.5cm} 
  \centering        
  \subfloat[Token Deletion]  
  {
      \label{fig:mtc_tokdel_valid}\includegraphics[width=0.32\textwidth]{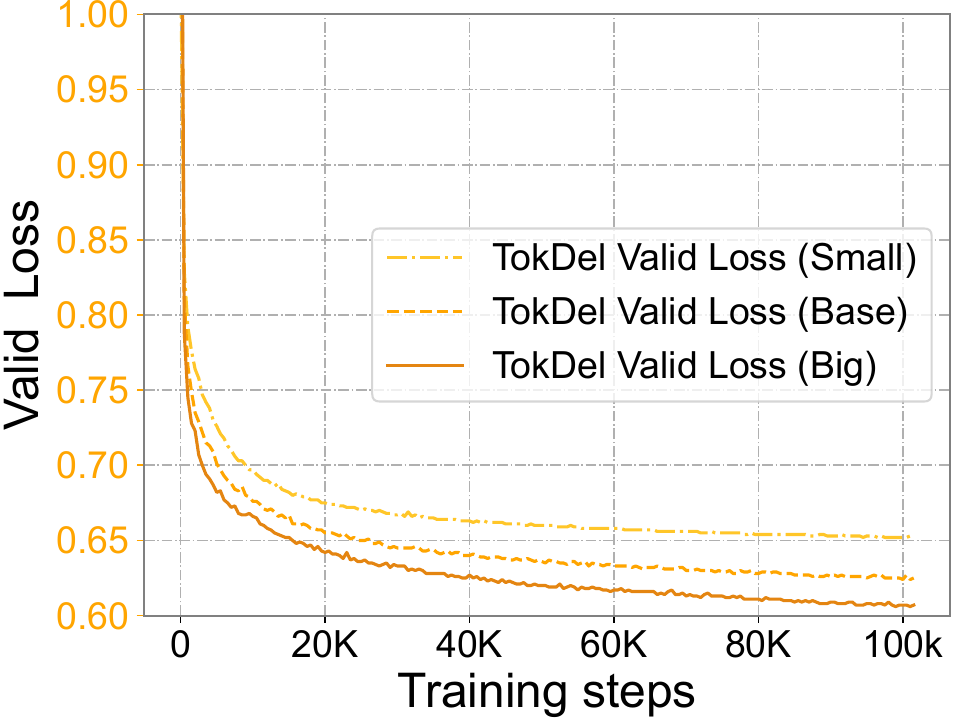}
  }
  \subfloat[Placeholder Insertion]
  {
      \label{fig:mtc_plhins_valid}\includegraphics[width=0.32\textwidth]{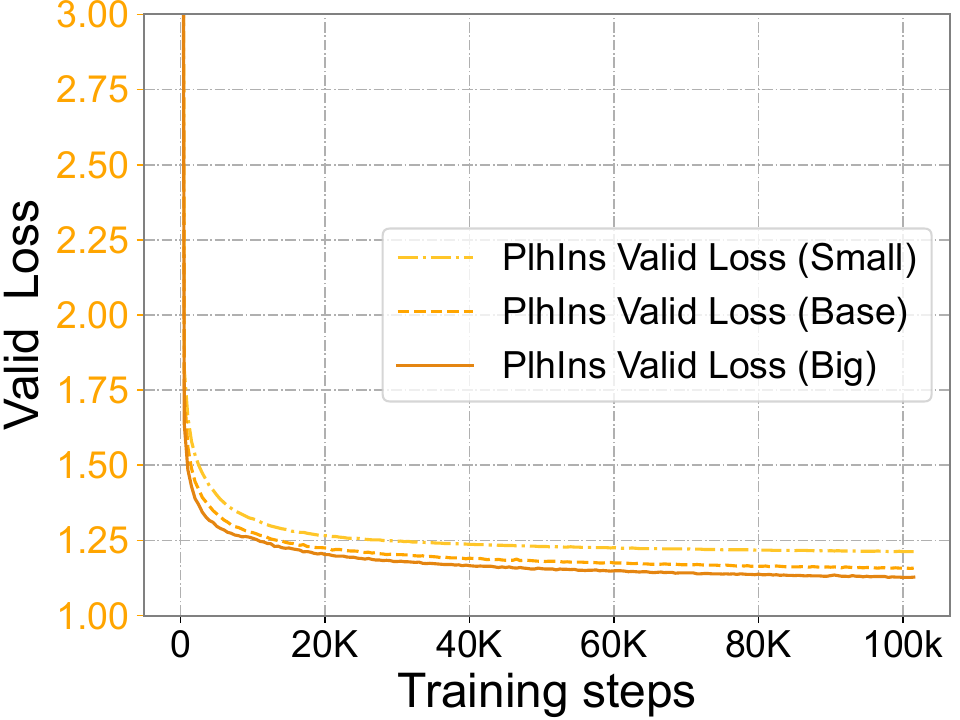}
  }
  \subfloat[Token Prediction]
  {
      \label{fig:mtc_tokpred_valid}\includegraphics[width=0.32\textwidth]{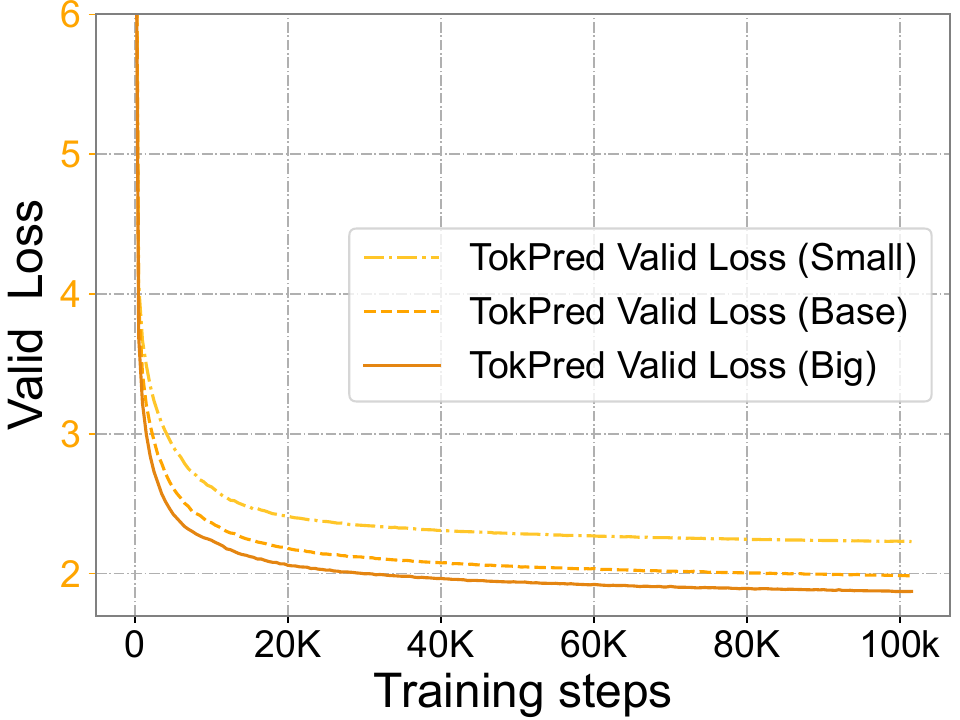}
  }

  \caption{\textbf{Validation Loss Curves of Editing Operations.} 
Similar to the training loss curves, the validation loss for the token prediction process shown in Figure (c) consistently remains the highest among the three types, while the loss for token deletion is the lowest. Additionally, as the model size increases, the validation loss for all three editing operations exhibits a stable downward trend.}    
  \label{fig:more_valid_curve_levt}            
\end{figure}

We present the changes in training and validation loss curves for \method{} models of varying sizes during training. As shown in Figure \ref{fig:more_train_curve_levt} and Figure \ref{fig:more_valid_curve_levt}, both training and validation losses for the three types of editing operations exhibit a stable downward trend as model scale increases. The loss from the token prediction process consistently constitutes the largest portion of the overall training loss. As is mentioned before, during the edit-based pre-training, the token prediction task is similar to that of MLM, as it involves predicting the real tokens corresponding to each placeholder token \texttt{[P]}, aiming to restore the complete target SMILES. However, unlike the results in Figure \ref{fig:saturation}, the token prediction loss in the \method{} pre-training does not exhibit a rapid saturation phenomenon in the early stages of training. Even in later training phases, the token prediction loss continues to decline steadily. This highlights the advantage of incorporating fragment-level training signals; by removing substructures and requiring the model to predict them, rather than randomly masking tokens, we establish a training task with improved scalability.

\section{Hyper-parameters for models of varying scales}
\label{more_pra_model}

In Table \ref{table_hyper_models}, we present the specific training hyperparameters for the models of different sizes (i.e., Big, Base, Small) used in this study. Notably, despite the differences in training objectives between MLM and \method{}, all other model settings, including the training hyperparameters (as listed in Table~\ref{table_hyper_models}) and training datasets, remain consistent to ensure a fair comparison of results.

\begin{table}[ht]
% \vspace{-0.3cm} 
\footnotesize
\color{black}{
\caption{Hyper-parameters for pre-train models with different scales.}
\label{table_hyper_models}

\begin{center}
\begin{tabular}{cccccccc}
\toprule
\multicolumn{1}{c}{\multirow{1}{*}{Model}} &\multicolumn{1}{c}{\multirow{1}{*}{Max Tokens}} &\multicolumn{1}{c}{\multirow{1}{*}{Layers}} &\multicolumn{1}{c}{\multirow{1}{*}{Attn Heads}} &\multicolumn{1}{c}{\multirow{1}{*}{Embed Dim}} &\multicolumn{1}{c}{\multirow{1}{*}{FFN Dim}} &\multicolumn{1}{c}{\multirow{1}{*}{Dropout}} 
&\multicolumn{1}{c}{\multirow{1}{*}{Num of Paras}} \\
\midrule

Big & 64K & 9 & 12 & 768 & 2048 & 0.1 &  50.5M \\
Base & 64K & 6 & 8 & 512 & 2048 & 0.1 & 19.4M \\
Small & 64K & 3 & 8 & 512 & 1024 & 0 & 6.8M\\
\bottomrule
\end{tabular}
\end{center}
}
% \vspace{-0.5cm} 
\end{table}

\section{Performance of MLMs on Downstream Task}
\label{mlm_down_step}

\begin{figure*}[ht]
% \vspace{-0.2cm}
% \vspace{-0.8cm}
\centering
	\includegraphics[width=0.72\linewidth]{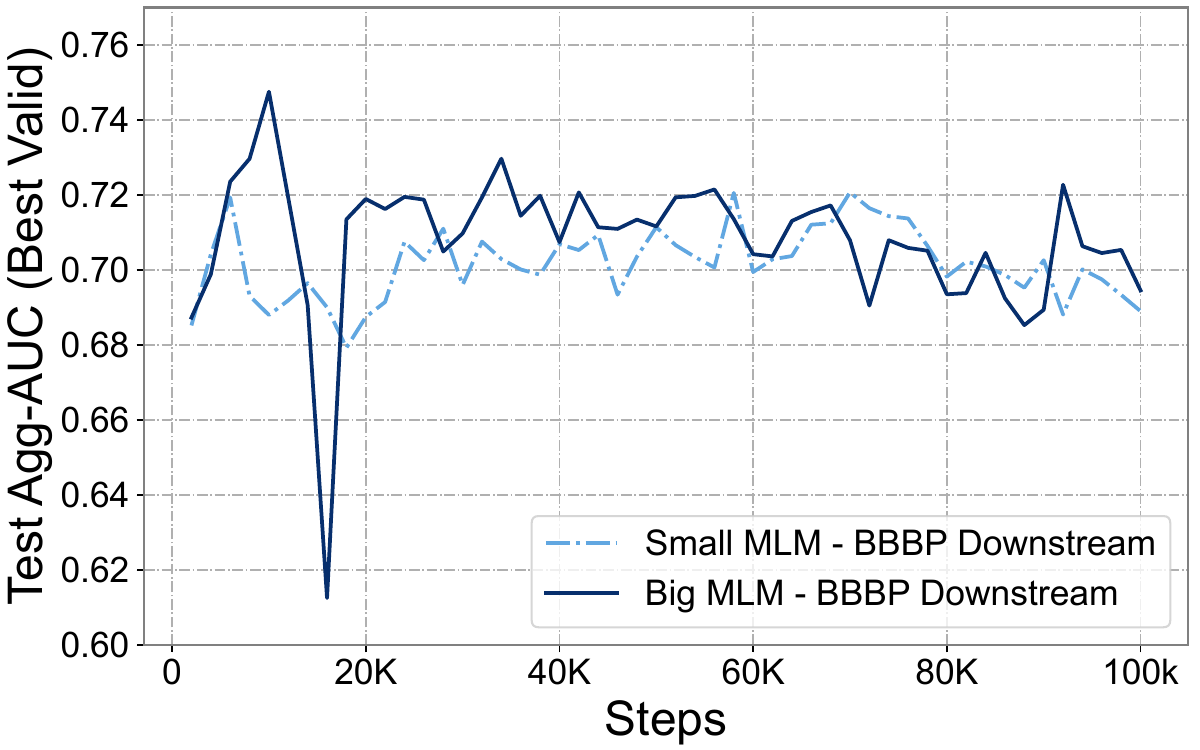}
 
	\caption {The performance of MLMs of different sizes and training steps on BBBP task.}
	\label{fig::mlm_downstream_bbbp}
% \vspace{-0.2cm}
\end{figure*}

We test the performance of MLMs (i.e., \mlmmth) of different sizes and different numbers of training steps on the BBBP task. To reduce variability and ensure accuracy, we evaluate each checkpoint on the downstream tasks five times and take the mean results. As shown in Figure \ref{fig::mlm_downstream_bbbp}, increasing the model's scale does not consistently improve the model's performance on the downstream tasks. In many cases, small models even outperform larger ones. This indicates that the semantic information captured by larger MLMs does not necessarily translate into better downstream task performance. Instead, larger models exhibit greater variability in their performance compared to small ones. These findings highlight the limited scalability and stability of MLMs.

\section{Performance of \method{} on Downstream Task}
\label{levt_down_step}

\begin{figure*}[ht]
% \vspace{-0.2cm}
% \vspace{-0.8cm}
\centering
	\includegraphics[width=0.72\linewidth]{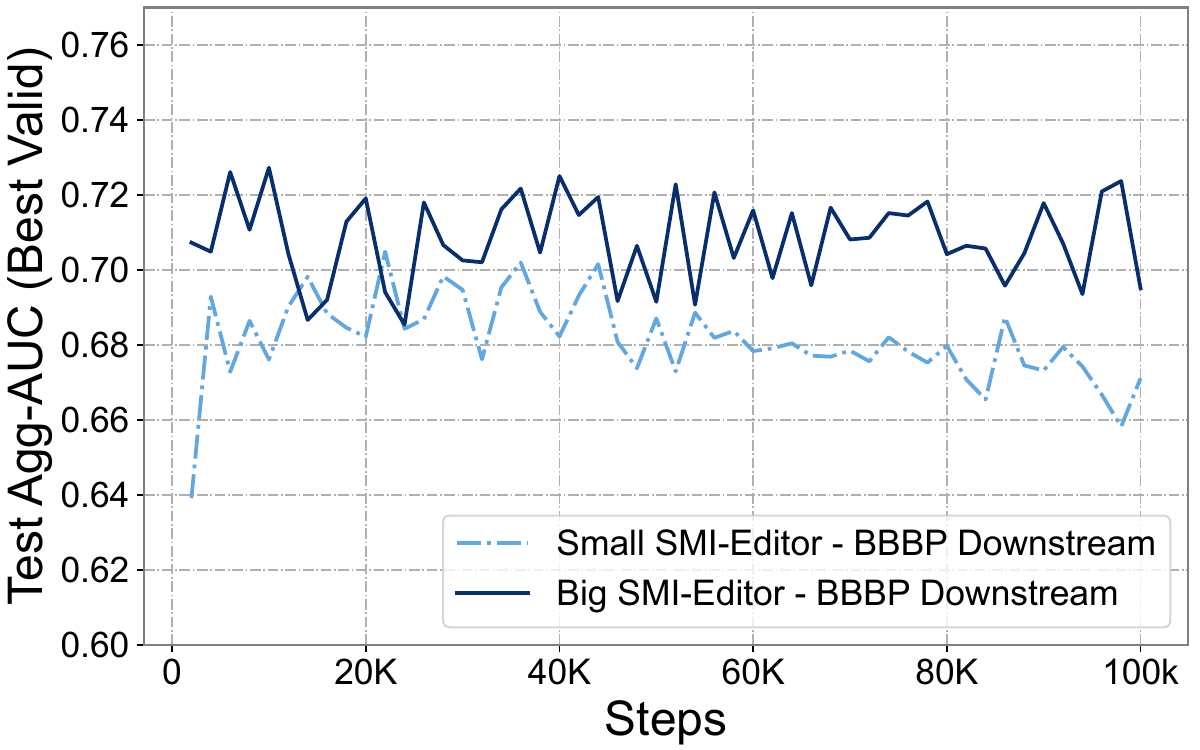}
 
	\caption {The performance of \method{} of different sizes and training steps on BBBP task.}
	\label{fig::levt_downstream_bbbp}
% \vspace{-0.2cm}
\end{figure*}

We also test the performance of \method{} of different sizes and training steps on the BBBP task, and the results are shown in Figure \ref{fig::levt_downstream_bbbp}. Again, we evaluate each checkpoint on the downstream tasks five times and take the mean result to ensure accuracy. Compared to the performance of the MLM (Figure \ref{fig::mlm_downstream_bbbp}), the larger \method{} model (i.e., Big Model) consistently outperforms the smaller models (i.e., Small Model). As the number of training steps increases, the performance gap between large and small models becomes more pronounced for \method{}. In contrast, MLMs do not exhibit this trend, with models of different sizes achieving similar performance on downstream tasks. Moreover, larger MLMs show greater performance fluctuations compared to their smaller counterparts. Notably, the larger \method{} model demonstrates superior performance stability, as it does not exhibit increased fluctuations compared to smaller \method{} models. These results indicate that {the \method{} model offers better training stability and model scalability than the MLM}.

\section{Hyper-Parameter Configuration for Pre-training}
\label{sec::pretrain_configuration}

We implement \method{} using $12$ stacked Transformer layers, each with $12$ attention heads. The model dimension and feedforward dimension of each Transformer layer are $768$ and $3{,}072$, respectively. The total number of \method{}'s parameters achieves $86.3$M. We use Adam \citep{kingma2014adam} optimizer and polynomial learning rate scheduler to train \method{}, and we set the learning rate as $5e-4$ and warmup step as $10$K.
The total training step is $120$K and each batch has $64$k tokens at maximum. We implement the \method{} model using the Fairseq library \footnote{\url{https://fairseq.readthedocs.io/en/latest/}} and train \method{} on four RTX3090 GPUs for about $1$ day.

For more pre-training hyper-parameters, please refer to Table \ref{table_hyper}.

\begin{table}[ht]
% \vspace{-0.3cm} 
\footnotesize
\color{black}{
\caption{\method{} hyper-parameters for pre-training.}
\label{table_hyper}

\begin{center}
\begin{tabular}{ccc}
\toprule
\multicolumn{1}{c}{\multirow{1}{*}{Hyper-parameters}} &\multicolumn{1}{c}{\multirow{1}{*}{Value}}\\
\midrule
Learning rate  & 5e-4 \\
LR scheduler  & polynomial\_decay \\
Warmup updates & 10K \\
Max updates & 120K \\
Max tokens & 64k \\
FFN dropout & 0.1 \\
Attention dropout & 0.1 \\
Activation dropout & 0 \\
Num of layers & 12 \\
Num of attention heads & 12 \\
Encoder embedding dim & 768 \\
Encoder FFN dim & 3072 \\
Adam ($\beta_1, \beta_2$) & (0.9,0.98) \\
Fragments Drop ratio &  0.15 \\
Vocabulary size & 369 \\
Activation function & GELU \\
Weight Decay & 0.0 \\
Clip Norm & 1.0 \\
\bottomrule
\end{tabular}
\end{center}
}
% \vspace{-0.5cm} 
\end{table}

\section{Hyper-Parameter Configuration for Fine-tuning}
\label{sec::ft_configuration}

In different downstream task, we use different hyper-parameters. For detailed fine-tuning hyper-parameters, please refer to Table \ref{table_hyper_ft}.
\begin{table}[ht]
% \vspace{-0.3cm} 
\footnotesize
\color{black}{
\caption{\method{} hyper-parameters for fine-tuning.}
\label{table_hyper_ft}

\begin{center}
\begin{tabular}{ccccccc}
\toprule
\multicolumn{1}{c}{\multirow{1}{*}{Tasks}} &\multicolumn{1}{c}{\multirow{1}{*}{Epochs}} &\multicolumn{1}{c}{\multirow{1}{*}{Batch size}} &\multicolumn{1}{c}{\multirow{1}{*}{Learning rate}} &\multicolumn{1}{c}{\multirow{1}{*}{Warmup Ratio}} &\multicolumn{1}{c}{\multirow{1}{*}{Dropout}} &\multicolumn{1}{c}{\multirow{1}{*}{Pooler-dropout}} \\
\midrule

BACE & 60 & 64 & 1e-4 & 0.06 & 0.1 & 0.2 \\
BBBP & 40 & 128 & 4e-4 & 0.06 & 0.1 & 0.1 \\
TOX21 & 80 & 128 & 1e-4 & 0.06 & 0.1 & 0.1 \\
SIDER & 100 & 32 & 5e-4 & 0.4 & 0.1 & 0 \\
MUV & 40 & 128 & 2e-5 & 0.2 & 0.1 & 0.1 \\
ClinTox & 100 & 256 & 5e-5 & 0.1 & 0.1 & 0.5 \\
ToxCast & 80 & 64 & 1e-4 & 0.06 & 0.1 & 0.1 \\
\bottomrule
\end{tabular}
\end{center}
}
% \vspace{-0.5cm} 
\end{table}

\section{Details of Fine-tuning Datasets}
\label{sec::ft_datasets}
We perform a comprehensive set of experiments on the MoleculeNet\citep{wu2018moleculenet} benchmark, focusing on the molecular property prediction task. MoleculeNet has emerged as one of the most widely recognized and utilized benchmarks in the field of molecular property prediction, providing a standardized platform for evaluating machine learning models' performances on evaluating molecular properties. Its datasets encompass a broad range of molecular tasks, and address diverse and practical scientific problems such as drug discovery, toxicity prediction and so on.

In this section, we provide a detailed summary of the statistics and fundamental characteristics of the MoleculeNet datasets we use in Table \ref{fine_tune_dataset}. This table offers information about the dataset sizes, task types, and compositions, providing readers with essential background information to better understand the experimental setup and subsequent analysis.

\begin{table*}[ht]
\caption{Summary information of the MoleculeNet benchmark datasets.\label{tab1}}
\tabcolsep=0pt%%
\begin{tabular*}{\textwidth}{@{\extracolsep{\fill}}ccccp{0.25\textwidth}@{\extracolsep{\fill}}}
\toprule
Dataset & Tasks & Task type & Molecules (train/valid/test) &  Describe \\
\midrule
ESOL &  1 & Regression  & 902/113/113 & Water solubility \\
FreeSolv &  1 & Regression & 513/64/64 & Hydrogen free energy \\
Lipo &  1 & Regression & 3,360/420/420 & Octanol/water distribution ratio, coefficient \\
BACE &  1 & Classification & 1,210/151/151 & Binding results of human BACE-1 inhibitors \\
BBBP &  1 & Classification  & 1,631/204/204 & Blood-brain barrier penetration \\
ClinTox &  2 & Multi-label classification  & 1,182/148/148 & Clinical trial toxicity and FDA approval status \\
Tox21 &  12 & Multi-label classification  & 6,264/783/783 & Qualitative toxicity measurements \\
ToxCast &  617 & Multi-label classification & 6,860/858/858 & Toxicology data based on in vitro screening \\
SIDER &  27 & Multi-label classification & 1,141/143/143 & Adverse drug reactions to the 27 systemic organs \\
MUV &  17 & Multi-label classification & 74,469/9,309/9,309 & A subset of PubChem BioAssay \\
\bottomrule
\end{tabular*}
\label{fine_tune_dataset}
\end{table*}

{\color{black}

\section{Performance of \method{} on DeepChem Data}
\label{sec::deepchem_down}

We evaluated the performance of \method{} (pre-trained on datasets provided by \citet{ross2022large}) on various downstream tasks of MoleculeNet benchmark using the data splits provided by DeepChem \footnote{https://github.com/deepchem/deepchem}. In our previous experiments, our results were based on a different data split, which made it less convincing to compare our model against others built on this dataset. Therefore, we re-tested \method{} on DeepChem splits and included comparisons with more baseline models. Detailed results are presented in Table~\ref{table_class_deep}. As shown in Table~\ref{table_class_deep}, {\method{} achieves significant performance gains over baseline models, reaching state-of-the-art levels with noticeable mean performance improvements}. Below is a detailed analysis of these results:

\begin{itemize}
    \item \method{} outperforms models trained with various paradigms: Measured by the mean performance, \method{} surpasses molecular representation learning models like MolCLR and ${\rm DMP_{TF}}$, which use contrastive pretraining, as well as models like ChemBerta and {\mlmmth}, which use masked language modeling. It also outperforms autoregressive LMs like Galactica and graph-based models like MolCLR, MGSSL, and MoMu. These results highlight the potential of SMILES LMs.
    \item \method{} achieves competitive performance with less training data: \method{} outperforms ${\rm DMP_{TF}}$, which is trained on over $100$ million compounds, despite using only $19$ million compounds for training. This demonstrates \method{}'s higher data efficiency, enabled by its ability to effectively leverage substructure information from SMILES sequences.
\end{itemize}

\begin{table}[ht]
% \vspace{-0.3cm} 
\footnotesize
\color{black}{
\caption{Mean results on MoleculeNet datasets using DeepChem splits. ROC-AUC scores (higher is better) are reported for all tasks. The best results are \textbf{bolded}}
\label{table_class_deep}
\begin{center}
\begin{tabular}{ccccccc|c}
\toprule
\multicolumn{1}{c}{\multirow{1}{*}{Method}} &\multicolumn{1}{c}{\multirow{1}{*}{BBBP↑}} &\multicolumn{1}{c}{\multirow{1}{*}{Tox21↑}} &\multicolumn{1}{c}{\multirow{1}{*}{ClinTox↑}} &\multicolumn{1}{c}{\multirow{1}{*}{HIV↑}} &\multicolumn{1}{c}{\multirow{1}{*}{BACE↑}} &\multicolumn{1}{c}{\multirow{1}{*}{SIDER↑}}&\multicolumn{1}{c}{\multirow{1}{*}{Mean↑}} \\
\midrule
GEM         & 72.4(0.4)    & 78.1(0.1)    & 90.1(1.3)     & 80.6(0.9)    & 85.6(1.1)    & 67.2(0.4)    & 79.0    \\ 
ChemBerta   & 64.3    & -       & 90.6     & 62.2    & -       & -       & -       \\ 
MolCLR      & 73.6(0.5)    & 79.8(0.7)    & 93.2(1.7)     & 80.6(1.1)    & 89.0(0.3)    & 68.0(1.1)    & 80.7    \\ 
MGSSL       & 70.5(1.1)    & 76.5(0.4)    & 80.7(2.2)     & 79.5(1.1)    & 79.7(0.8)    & 61.8(0.7)    & 74.8    \\ 
${\rm DMP_{TF}}$ & 78.1(0.5)    & 78.8(0.5)    & 95.0(0.5)     & 81.0(0.7)    & 89.3(0.9)    & 69.2(0.7)    & 81.9    \\ 
% Galactica   & 66.1    & 68.9    & 82.6     & 74.5    & 61.7    & 63.2    & 69.5    \\ 
MoMu        & 70.5(2.0)    & 75.6(0.3)    & 79.9(4.1)     & 76.2(0.9)    & 77.1(1.4)    & 60.5(0.9)    & 73.3    \\ 
\hline
\mlmmth     & 89.4(1.9)    & 76.2(1.6)    & 90.6(1.8)     & 79.8(1.2)    & 86.6(0.4)    & 66.5(0.5)    & 81.5    \\ 
\method{}  & \textbf{93.5}(2.2) & \textbf{81.4}(1.1) & \textbf{95.2}(1.3) & \textbf{81.6}(0.7) & \textbf{89.9}(0.2) & \textbf{69.8}(0.6) & \textbf{85.2} \\ 
\bottomrule
\end{tabular}
\end{center}
} 
\end{table}

\section{Performance Advantages of \method{} Over Auto-regressive Models}
\label{sec::gpt_cmp}

To comprehensively compare \method{} with autoregressive models, we trained a decoder-only model with identical architecture and size to \method{} using an autoregressive language-modeling objective, referred to as SMI-GPT. We evaluated SMI-GPT's performance across several downstream tasks, with results shown in Table \ref{table_smi_gpt}. The findings indicate that \method{} can perform better than SMI-GPT. Below is an analysis of these results:

\begin{table}[ht]
% \vspace{-0.3cm} 
\footnotesize
\color{black}{
\caption{Results of \method{} and SMI-GPT on MoleculeNet datasets using DeepChem splits. ROC-AUC scores (higher is better) are reported for all tasks The best results are \textbf{bolded}}
\label{table_smi_gpt}
\begin{center}
\begin{tabular}{ccccccc|c}
\toprule
\multicolumn{1}{c}{\multirow{1}{*}{Method}} &\multicolumn{1}{c}{\multirow{1}{*}{BBBP↑}} &\multicolumn{1}{c}{\multirow{1}{*}{Tox21↑}} &\multicolumn{1}{c}{\multirow{1}{*}{ClinTox↑}} &\multicolumn{1}{c}{\multirow{1}{*}{HIV↑}} &\multicolumn{1}{c}{\multirow{1}{*}{BACE↑}} &\multicolumn{1}{c}{\multirow{1}{*}{SIDER↑}}&\multicolumn{1}{c}{\multirow{1}{*}{Mean↑}} \\
\midrule

SMI-GPT(NT)&88.5&74.3&88.9&68.8&76.2&63.7&76.7    \\ 
SMI-GPT(Emb)&91.2&75.1&91.4&79.4&86.2&67.1&81.7 \\
MoMu        & 70.5    & 75.6    & 79.9     & 76.2    & 77.1    & 60.5    & 73.3    \\ 
\mlmmth     & 89.4    & 76.2    & 90.6     & 79.8    & 86.6    & 66.5    & 81.5    \\ 
\method{}  & \textbf{93.5} & \textbf{81.4} & \textbf{95.2} & \textbf{81.6} & \textbf{89.9} & \textbf{69.8} & \textbf{85.2} \\ 
\bottomrule
\end{tabular}
\end{center}
} 
\end{table}

\subsection{Implementation details for SMI-GPT(NT) and SMI-GPT(Emb)}

\textbf{ SMI-GPT(NT)}: This approach uses next-token prediction for downstream classification tasks by appending a special token (e.g., $Label_{0}$, $Label_{1}$) at the end of each SMILES sequence to denote the classes of sample's label. The model learns to predict the correct label token during fine-tuning.
\textbf{SMI-GPT(Emb)}: The representations of each token in the SMILES string extracted by the SMI-GPT model are processed using mean pooling. The resulting pooled representation is then fed into a classification head, which predicts the class of the SMILES. 

\subsection{Advantages of the encoder-only \method{} architecture}

As shown in Table~\ref{table_smi_gpt}, \method{} consistently outperforms SMI-GPT(Emb) and SMI-GPT(NT), highlighting its superior semantic learning capabilities.  SMI-GPT(Emb) achieves better performance than SMI-GPT(NT), suggesting that pretraining-based feature transfer is preferable for molecular property prediction tasks. Therefore, the encoder-only pre-trained model is highly suitable for molecular property prediction tasks.

\label{analyse_gpt_curve}
\begin{figure}[t]    
  \centering          
  \subfloat[Train Loss]  
  {
      \label{fig:gpt_subfig1}\includegraphics[width=0.47\textwidth]{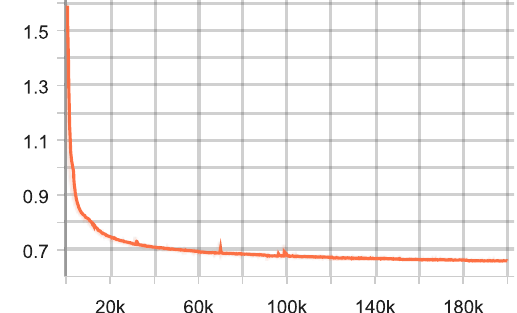}
  }
  \subfloat[Train PPL]
  {
      \label{fig:gpt_subfig2}\includegraphics[width=0.47\textwidth]{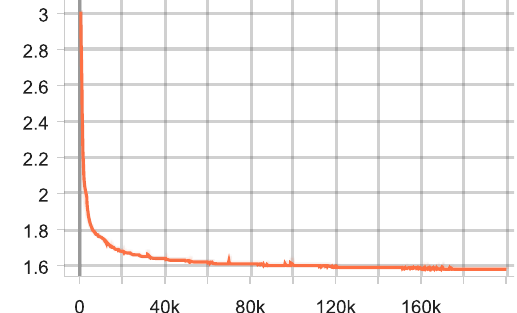}
  }

  \caption{The training loss and perplexity (PPL) curves of the SMI-GPT model.}   
  \label{fig:get_curve}            
\end{figure}

\subsection{Rapid convergence in autoregressive LMs}

We provide the training curve of the SMI-GPT model in Figure \ref{fig:get_curve}, which shows that the loss decreases rapidly during the early stages of training. Similarly, the perplexity also drops quickly, reaching approximately $1.6$ at the $40$K training step. By the end of training, the model's Perplexity falls below $1.6$, which is significantly lower than the perplexity typically observed for GPT models trained on text data.

\subsection{Why does this phenomenon occur?}

For auto-regressive LMs, each time a new token is generated, it receives all preceding tokens as prefix input. This means that when the model generates tokens at later positions, it has access to more comprehensive contextual information (i.e., a longer prefix and more complete sequence information). As a result, {the prediction difficulty for tokens in later positions is significantly reduced, allowing the model to converge more easily}. A key difference between \method{} and SMI-GPT is that in \method{}, each discarded token is predicted independently, with equal importance assigned to the prediction of each token. This enables \method{} to more effectively capture the complete semantic information encoded in the tokens.

In summary, compared to LLMs on text data, GPT models on SMILES data converge significantly faster and achieve much lower perplexity. This indicates that SMILES data is inherently easier to fit than natural-language text. Therefore, it is crucial to design effective methods to extract richer semantic information from SMILES. Our {\method{} represents a meaningful and successful exploration in this direction, highlighting the importance of leveraging substructural fragment information within SMILES data.}

\section{Performance of \method{} with Fragment Correction}
\label{sec::smi_cor}
\subsection{Training \method{} to correct errors and remove extraneous components did not improve performance} 
We implemented a version of \method{} that learns to correct erroneous functional groups and remove extraneous substructures, referred to as \method{}-Cor. However, \method{}-Cor did not outperform the original \method{} on downstream tasks. Table \ref{table_smi_cor} below compares the performance of \method{} and \method{}-Cor, showing that their performance is similar, demonstrating the limited benefit of incorporating these tasks.

\subsection{Analysis of \method{}-Cor’s performance}
We attribute \method{}-Cor's lack of improvement to the following reasons:  
\begin{itemize}
    \item \textbf{Correcting errors and removing extraneous components provide limited additional training signals:}  
     \method{}’s training comprises two major steps: deletion and insertion. During deletion, erroneous functional groups and extraneous substructures are removed, while the insertion step involves learning to recover the correct tokens in the appropriate positions. Thus adding erroneous functional groups or extraneous substructures affects only the deletion step, which is a simpler task providing limited information. Moreover, as shown in Table \ref{table::ablation_edit} of the main text, ablating the token deletion (TokDel) step has minimal performance impact.  
     \item \textbf{Identifying erroneous functional groups and extraneous structures is too simple for the model: } 
     \method{}-Cor constructs erroneous inputs through random substitutions, often resulting in chemically invalid SMILES that are easy for the model to identify. Consequently, the simplicity of the training task limits further performance improvement.
\end{itemize}

\begin{table}[ht]
% \vspace{-0.3cm} 
\footnotesize
\color{black}{
\caption{Performance comparison between \method{}-Cor and \method{}.}
\label{table_smi_cor}
\begin{center}
\begin{tabular}{cccccc|c}
\toprule
\multicolumn{1}{c}{\multirow{1}{*}{Method}} &\multicolumn{1}{c}{\multirow{1}{*}{BACE↑}} &\multicolumn{1}{c}{\multirow{1}{*}{BBBP↑}} &\multicolumn{1}{c}{\multirow{1}{*}{SIDER↑}} &\multicolumn{1}{c}{\multirow{1}{*}{Tox21↑}} &\multicolumn{1}{c}{\multirow{1}{*}{ToxCast↑}} &\multicolumn{1}{c}{\multirow{1}{*}{Mean↑}} \\
\midrule

\method{}-Cor        &80.6  & 77.1  & 62.2   & 76.8   & 68.0     & 72.9        \\ 
\method{}     & 80.3  & 77.4  & 63.0   & 77.1   & 67.4     & 73.0        \\ 
\bottomrule
\end{tabular}
\end{center}
} 
\end{table}

\section{How the fragment drop ratio affect \method{}}
\label{sec::drop_ratio}

To investigate the impact of the fragment drop ratio on \method{}, we trained \method{} models with different drop ratios (i.e., $15\%$, $30\%$, $45\%$) and analyzed their training curves and downstream task performance. The results indicate that increasing the drop ratio significantly raises training loss for \method{}, indicating that its pretraining task is more challenging than MLM. Below are the detailed findings:

\label{analyse_gpt_curve_dropratio}
\begin{figure}[t]    
  \centering          
  \subfloat[Train Loss]  
  {
      \label{fig:gpt_subfig1_dropratio}\includegraphics[width=0.47\textwidth]{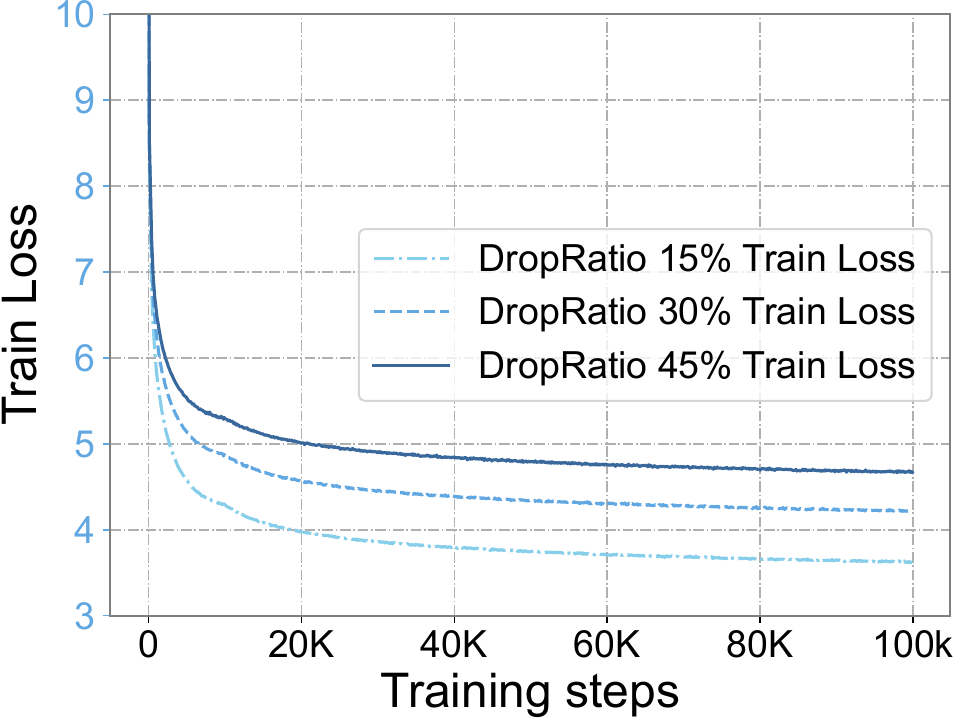}
  }
  \subfloat[Valid Loss]
  {
      \label{fig:gpt_subfig2_dropratio}\includegraphics[width=0.47\textwidth]{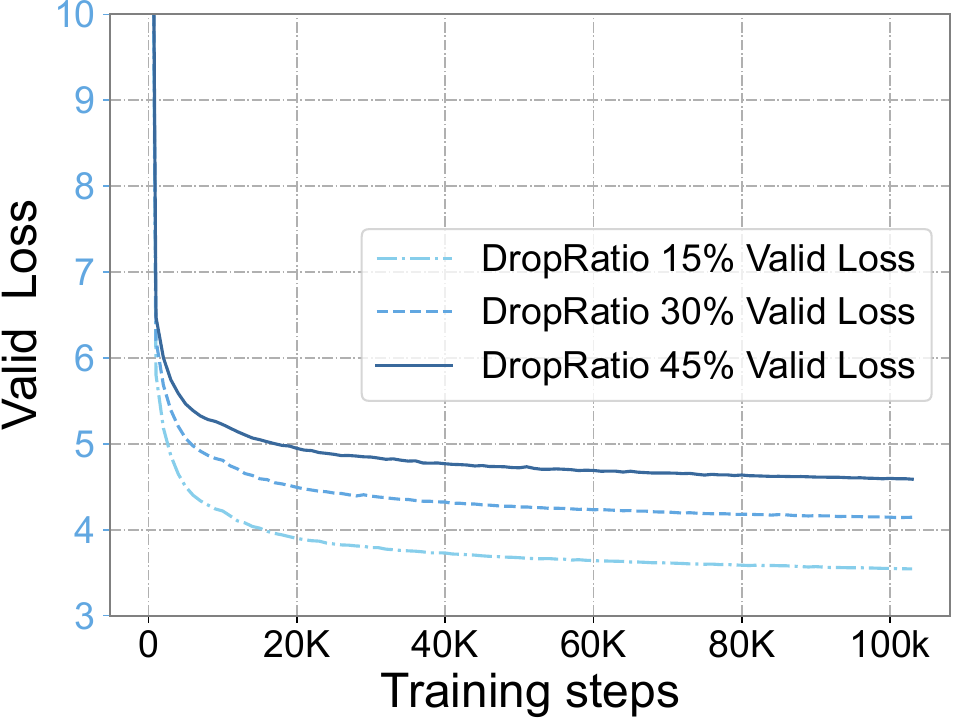}
  }

  \caption{The training loss and valid loss of the \method{} with different fragment drop ratios.}   
  \label{fig:drop_ratio_edit}            
\end{figure}

\subsection{Impact on \method{}'s convergence} 

We plotted the training and validation loss curves for \method{} with varying drop ratios in Figure \ref{fig:drop_ratio_edit}. \textbf{The results show that as the drop ratio increases, both training and validation losses rise significantly.} Compared to Figure \ref{fig:sat_subfig3} of the paper, the loss increase for \method{} is more pronounced than for MLM, confirming that \method{}'s task is inherently more challenging.

\subsection{Impact on downstream task performance} 
We evaluated the performance of \method{} and MLMs with varying drop or mask ratios. The results are summarized in Table \ref{table_drop_ratio}. From Table \ref{table_drop_ratio}, it can be observed that as the mask ratio increases, the average performance of the \mlmmth model shows no significant change, while the performance of the \method{} model declines as the drop ratio increases. This indicates that \method{} represents a more challenging training task.

\textbf{Here is a more detailed explanation: } 
\begin{itemize}
    \item \method{} discards chemically-meaningful substructures that often serve as standalone semantic units. This also makes predicting the discarded fragments more difficult than predicting individual masked tokens. Dropping more substructures severely disrupts the molecular structure, making it harder for the model to reconstruct the original molecule.
    \item  MLM, on the other hand, randomly masks tokens in SMILES sequences. Since SMILES tokens often represent individual atoms or bonds, masking does not typically disrupt the molecular semantics significantly. For instance, masking one or two atoms of a functional group like ${\rm -COOH}$ still leaves enough contextual information to reconstruct it. Additionally, the probability of masking an entire functional group is low due to MLM's token-based masking mechanism. This explains why MLM performance is less sensitive to mask ratio increases, as also reflected in Figure \ref{fig:sat_subfig3} of the paper: Different Mask Ratios Cannot Alleviate Rapid Saturation.
\end{itemize}

\begin{table}[ht]
\footnotesize
\color{black}{
\caption{Performance of \method{} and \mlmmth with different drop or mask ratios on downstream tasks.}
\label{table_drop_ratio}
\begin{center}
\begin{tabular}{cccccc|c}
\toprule
\multicolumn{1}{c}{\multirow{1}{*}{Method}} &\multicolumn{1}{c}{\multirow{1}{*}{BACE↑}} &\multicolumn{1}{c}{\multirow{1}{*}{BBBP↑}} &\multicolumn{1}{c}{\multirow{1}{*}{SIDER↑}} &\multicolumn{1}{c}{\multirow{1}{*}{Tox21↑}} &\multicolumn{1}{c}{\multirow{1}{*}{ToxCast↑}} &\multicolumn{1}{c}{\multirow{1}{*}{Mean↑}} \\
\midrule

\mlmmth(15\%)        &77.8    & 68.6  & 61.2  & 75.1   & 64.9   &69.5        \\ 
\mlmmth(30\%)        &78.3     & 70.2     & 58.2     &76.0    &63.7    & 69.3        \\ 
\mlmmth(45\%)       &78.4    & 66.1   & 59.3   & 76.4   & 65.5   & 69.1        \\ 
\method{}(15\%)     & 80.3    & \textbf{77.4}& \textbf{63.0} & 77.1     & \textbf{67.4} & \textbf{73.0}        \\ 
\method{}(30\%)     & \textbf{81.6} & 73.3    &59.6   &77.0   & 66.8   & 71.7       \\ 
\method{}(45\%)     & 79.3    & 72.2   & 61.1  & \textbf{77.8} & 67.1   & 71.5        \\ 
\bottomrule
\end{tabular}
\end{center}
} 
\end{table}

\section{Performance of \method{} on molecular property regression tasks}
\label{sec::smi_regress}

We evaluated the model's performance on three molecular property regression tasks, as shown in Table \ref{table_regre}. \method{} achieved the best performance compared to baseline models and significantly outperformed the MLM model.

\begin{table}[ht]
% \vspace{-0.3cm} 
\footnotesize
\color{black}{
\caption{Performance of \method{} on molecular property regression tasks.}
\label{table_regre}
\begin{center}
\begin{tabular}{cccc}
\toprule
\multicolumn{1}{c}{\multirow{1}{*}{Method}} &\multicolumn{1}{c}{\multirow{1}{*}{ESOL↓}} &\multicolumn{1}{c}{\multirow{1}{*}{FreeSolv↓}} &\multicolumn{1}{c}{\multirow{1}{*}{Lipo↓}}  \\
\midrule

MPNN       & 0.58      & 1.150     & 0.7190    \\
${\rm DMP_{TF}}$   & 0.700     & -         & -         \\
 A-FP       & 0.503     & 0.736     &0.578     \\
 \mlmmth    & 0.576     & 0.709     & 0.642     \\
 \method{} & \textbf{0.362} & \textbf{0.524} & \textbf{0.565} \\

\bottomrule
\end{tabular}
\end{center}
} 
\end{table}

\section{A case study for fragments assemble}
\label{sec::case_assem}
We provide an example workflow of data processing in pre-training with Paracetamol (SMILES: CC(=O)Nc1ccc(O)cc1) in Figure \ref{fig::case_assem}.
\begin{figure*}[t]

\centering
	\includegraphics[width=0.8\linewidth]{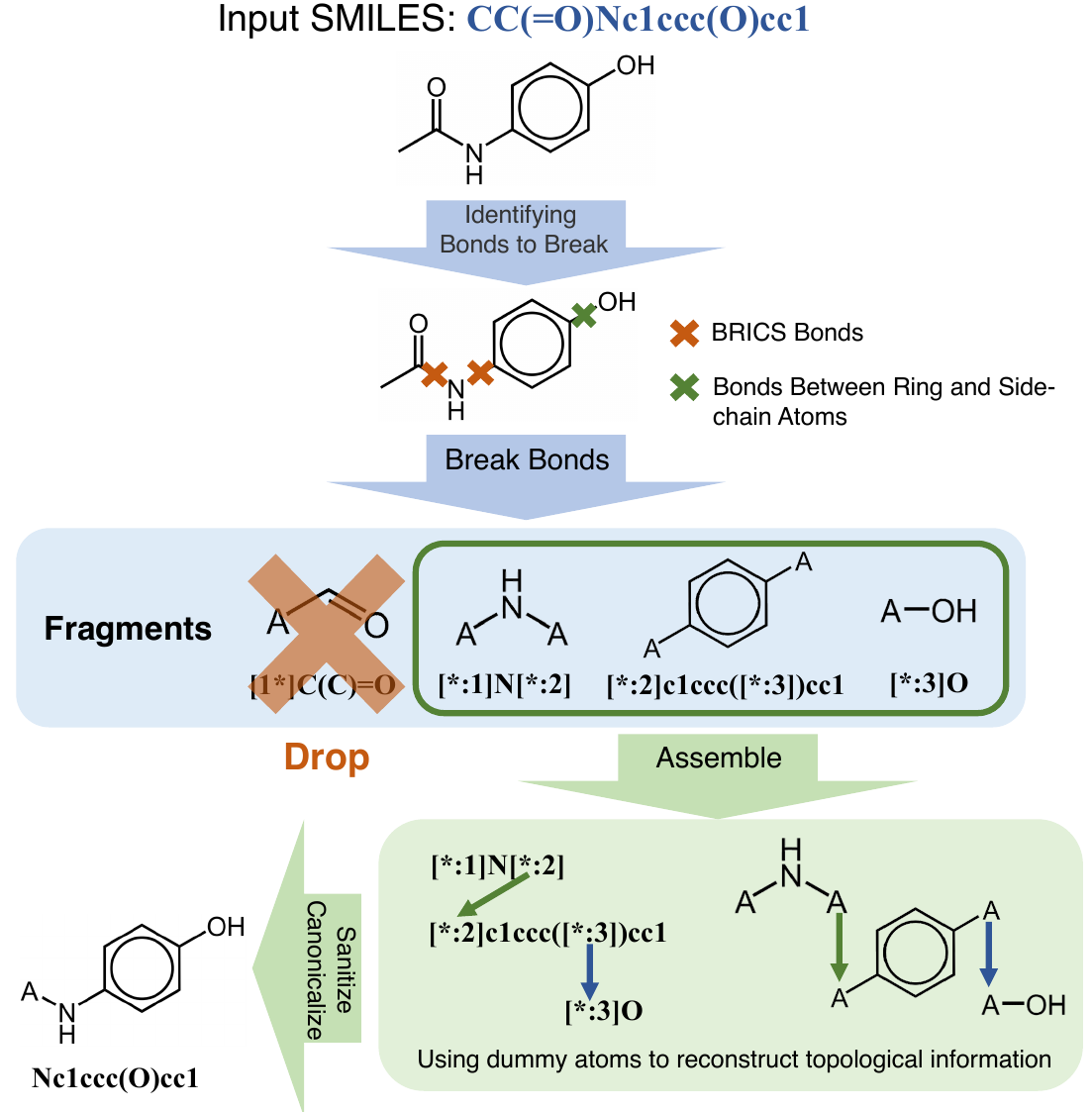}
 
	\caption {An Example Workflow of Molecule Fragmentation and Assemble with Paracetamol}
	\label{fig::case_assem}
\end{figure*}

\section{The scalability of \method{}}
\label{sec::scalability}

We added results showing the performance of \method{} and {\mlmmth} models of varying sizes on downstream tasks, which further demonstrate \method{}’s strong scalability. These results are shown in Table \ref{table_sacle}. It is evident that while increasing model size has minimal impact on MLMs, larger \method{} models show more consistent performance gains. \textbf{This confirms the claim that \method{} has better scalability compared to MLMs.}

\begin{table}[ht]
\footnotesize
\color{black}{
\caption{Performance of \method{} and \mlmmth with different scales on downstream tasks.}
\label{table_sacle}
\begin{center}
\begin{tabular}{cccccc|c}
\toprule
\multicolumn{1}{c}{\multirow{1}{*}{Method}} &\multicolumn{1}{c}{\multirow{1}{*}{BACE↑}} &\multicolumn{1}{c}{\multirow{1}{*}{BBBP↑}} &\multicolumn{1}{c}{\multirow{1}{*}{SIDER↑}} &\multicolumn{1}{c}{\multirow{1}{*}{Tox21↑}} &\multicolumn{1}{c}{\multirow{1}{*}{ToxCast↑}} &\multicolumn{1}{c}{\multirow{1}{*}{Mean↑}} \\
\midrule

\mlmmth(Small)        & 76.8     & 69.6     & 60.5     & 75.3     & 64.2     & 69.2            \\ 
\mlmmth(Base)        & 76.6     & 69.3     & 59.9     & 75.3     & 64.4     & 69.1        \\ 
\mlmmth(Big)       & 77.4     & 68.7     & 60.8     & 75.1     & 65.3     & 69.4         \\ 
\method{}(Small)     & 78.3     & 72.6     & 59.4     & 75.6     & 65.1     & 70.2          \\ 
\method{}(Base)     & 79.2     & 73.2     & \textbf{61.0} & 75.7     & 65.8     & 71.0       \\ 
\method{}(Big)     & \textbf{79.3} & \textbf{74.2} & 60.9     & \textbf{76.7} & \textbf{66.4} & \textbf{71.5}         \\ 
\bottomrule
\end{tabular}
\end{center}
} 
\end{table}

\section{The training cost of \method{}}
\label{sec::traing_cost}

We measured that the training cost of \method{} is approximately three times that of MLMs (i.e., \mlmmth) for the same model size, training hyperparameters, and data. However, {the training cost of \method{} remains acceptable}. To better analyze the impact of training cost, we trained an MLM with equivalent computational cost (\mlmmth(More)). Results showed that \mlmmth(More) performed worse than the original \mlmmth and significantly lagged behind \method{}, highlighting that merely increasing MLM training cost does not yield better results.

\subsection{Reasons for higher training cost in \method{}} \method{} requires computing expert actions (using a computationally expensive dynamic programming algorithm) and modeling three different editing operations, which introduces additional overhead.

\subsection{Acceptable training cost} 
Training \method{} on a dataset with $19$M compounds using four RTX 3090 GPUs took approximately $24.6$ hours. Scaling \method{} to larger datasets (e.g., more than $100$M compounds) is feasible, demonstrating its potential for broader applications.

\subsection{\method{} performs better under the same training cost with MLM}  
We trained \mlmmth(More) with the same computational cost as \method{} by increasing its training steps from $120$K to $360$K. Table \ref{table_mlm_more} shows that \mlmmth(More) performs worse than the \method{} and  original \mlmmth . This is due to rapid saturation issues in MLM training on SMILES data. \textbf{This also indicates that the speed of model training is not the most important factor; what matters more is whether the model can efficiently extract high-quality semantic representations}. This highlights the importance of designing more powerful training schemes like \method{} to effectively extract meaningful information from SMILES.

\subsection{Higher performance ceiling for \method{}} 
Although the inclusion of Experts slows down the training speed of the \method{} model, it also enriches the semantic information the model learns. This gives \method{} greater scalability and a higher performance ceiling compared to {\mlmmth}. As shown in Table \ref{table_mlm_more}, \method{} benefits more from more computational resources. This makes \method{} a better choice when given sufficient training budget.

\begin{table}[ht]
\footnotesize
\color{black}{
\caption{Performance of \method{} and {\mlmmth} with training steps on downstream tasks.}
\label{table_mlm_more}
\begin{center}
\begin{tabular}{cccccc|c}
\toprule
\multicolumn{1}{c}{\multirow{1}{*}{Method}} &\multicolumn{1}{c}{\multirow{1}{*}{BACE↑}} &\multicolumn{1}{c}{\multirow{1}{*}{BBBP↑}} &\multicolumn{1}{c}{\multirow{1}{*}{SIDER↑}} &\multicolumn{1}{c}{\multirow{1}{*}{Tox21↑}} &\multicolumn{1}{c}{\multirow{1}{*}{ToxCast↑}} &\multicolumn{1}{c}{\multirow{1}{*}{Mean↑}} \\
\midrule

\mlmmth(More)        & 74.3     & 66.2     & 49.5     & 73.3     & 62.3     & 65.1            \\ 
\mlmmth       & 77.8     & 68.6     & 61.2     & 75.1     & 64.9     & 69.5         \\ 
\method{}     & \textbf{80.3} & \textbf{77.4} & \textbf{63.0} & \textbf{77.1} & \textbf{67.4} & \textbf{73.0 }        \\ 
\bottomrule
\end{tabular}
\end{center}
} 
\end{table}

\section{A more detailed analysis of the model's substructure modeling capability}
\label{sec::detail_struct}

The observed trends for the FreeSolv dataset are fully consistent with our expectations and align with the definition of its physical properties. 
On the other hand, the performance on the ESOL dataset is influenced by additional factors such as molecular weight. We also designed more analytical experiments to further investigate the behavior of the \method{} model, and the results demonstrate that the model’s behavior aligns with expectations. Detailed explanations are as follows.

\begin{figure*}[t]

\centering
	\includegraphics[width=0.6\linewidth]{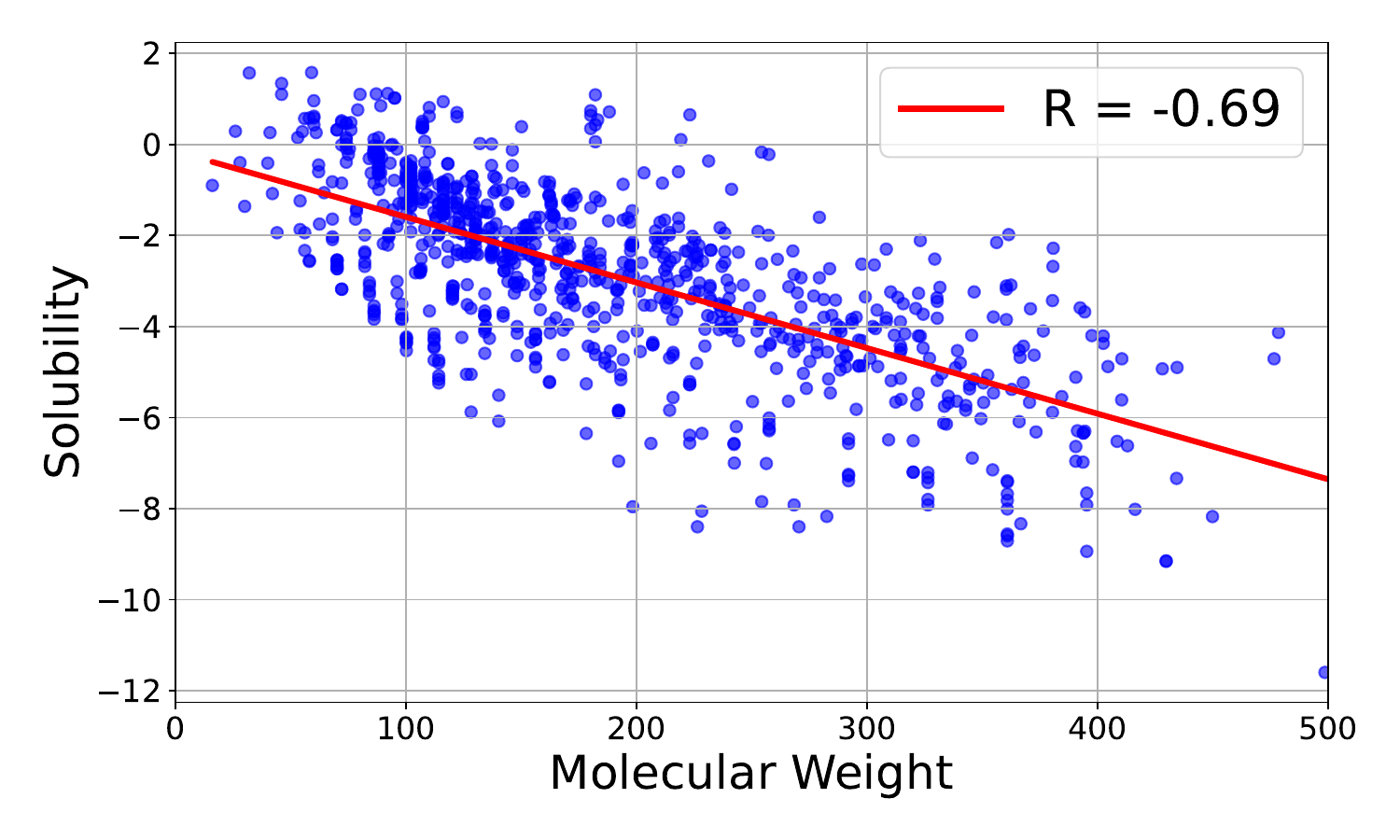}
 
	\caption {The relationship between molecular weight and solubility in the ESOL training set}
	\label{fig::esol_vis}
\end{figure*}

{For the FreeSolv dataset, the observed trends align with its physical property definitions.} 
FreeSolv reflects the hydration free energy of compounds, defined as the free energy change when a compound transitions from a non-dissolved state to a dissolved state. When hydrophilic groups in a molecule are reduced, the change in hydration free energy increases, leading to higher hydration free energy. Therefore, when we remove hydrophilic groups from the molecule, the model predicts an increase in hydration free energy, consistent with the trend observed in Figure \ref{fig:substruct_subfig2_levt}, which matches our expectations.

{For the ESOL task, the model predictions are significantly influenced by molecular weight} 
The ESOL dataset reflects compound solubility, which is strongly negatively correlated with molecular weight: the larger the molecular weight, the lower the solubility. We plotted a scatter diagram (Figure \ref{fig::esol_vis}) showing the relationship between molecular weight and solubility in the ESOL training set. A clear negative correlation with a coefficient of \( R = -0.69 \) is observed. Consequently, when functional groups or atoms are removed from a molecule, its molecular weight decreases, leading the model to predict an increase in solubility. This explains why, in Figure \ref{fig:substruct_subfig1_levt}, the model predicts increased solubility regardless of whether hydrophilic groups or random groups are removed. The increase is more significant with random deletions, demonstrating the model's ability to distinguish between hydrophilic group deletions and random deletions.

\begin{figure}[t]    
  \centering          
  \subfloat[\method{}]  
  {
      \label{fig:substruct_rep_subfig1}\includegraphics[width=0.49\textwidth]{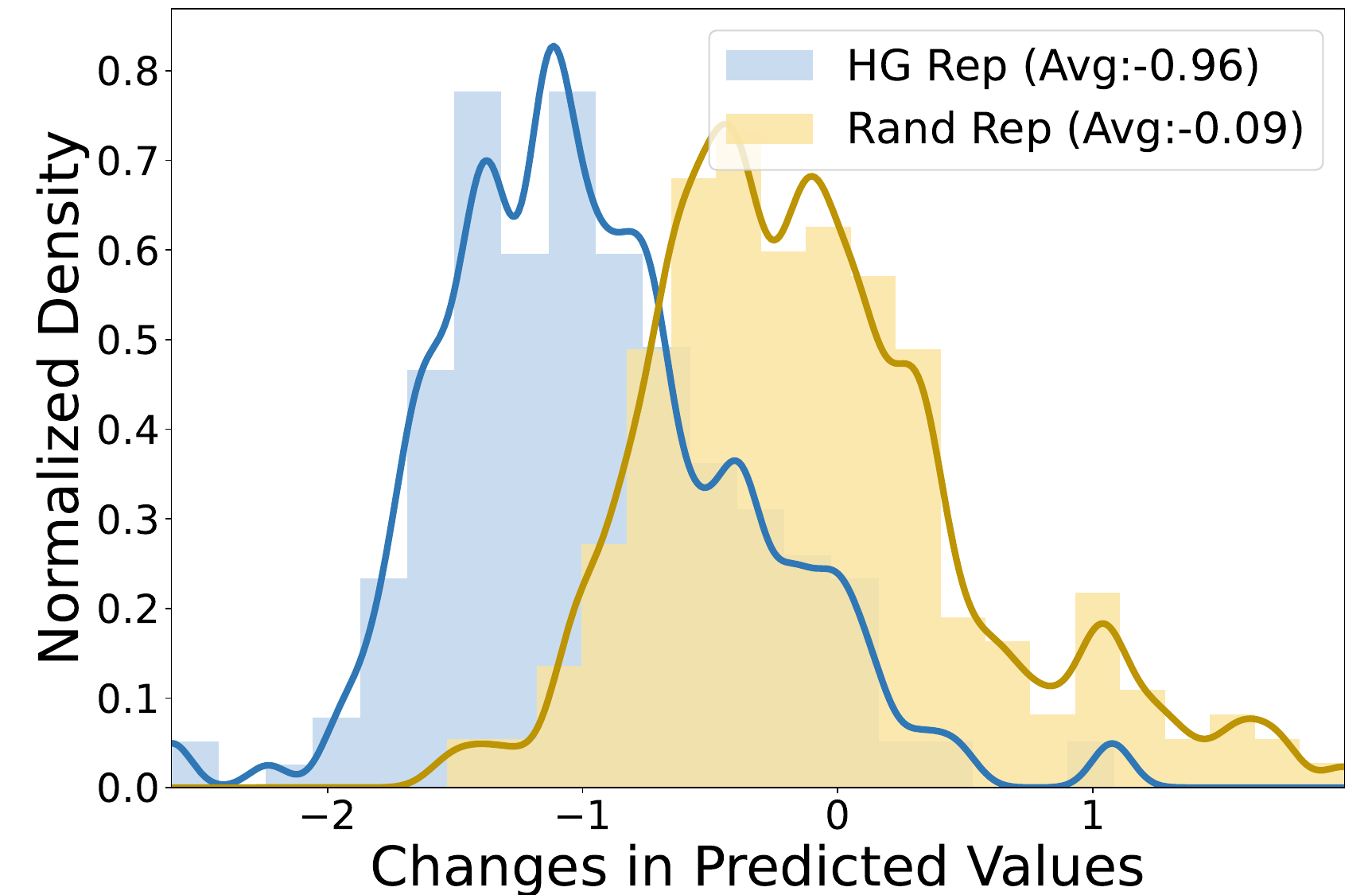}
  }
  \subfloat[\mlmmth]
  {
      \label{fig:substruct_rep_subfig2}\includegraphics[width=0.49\textwidth]{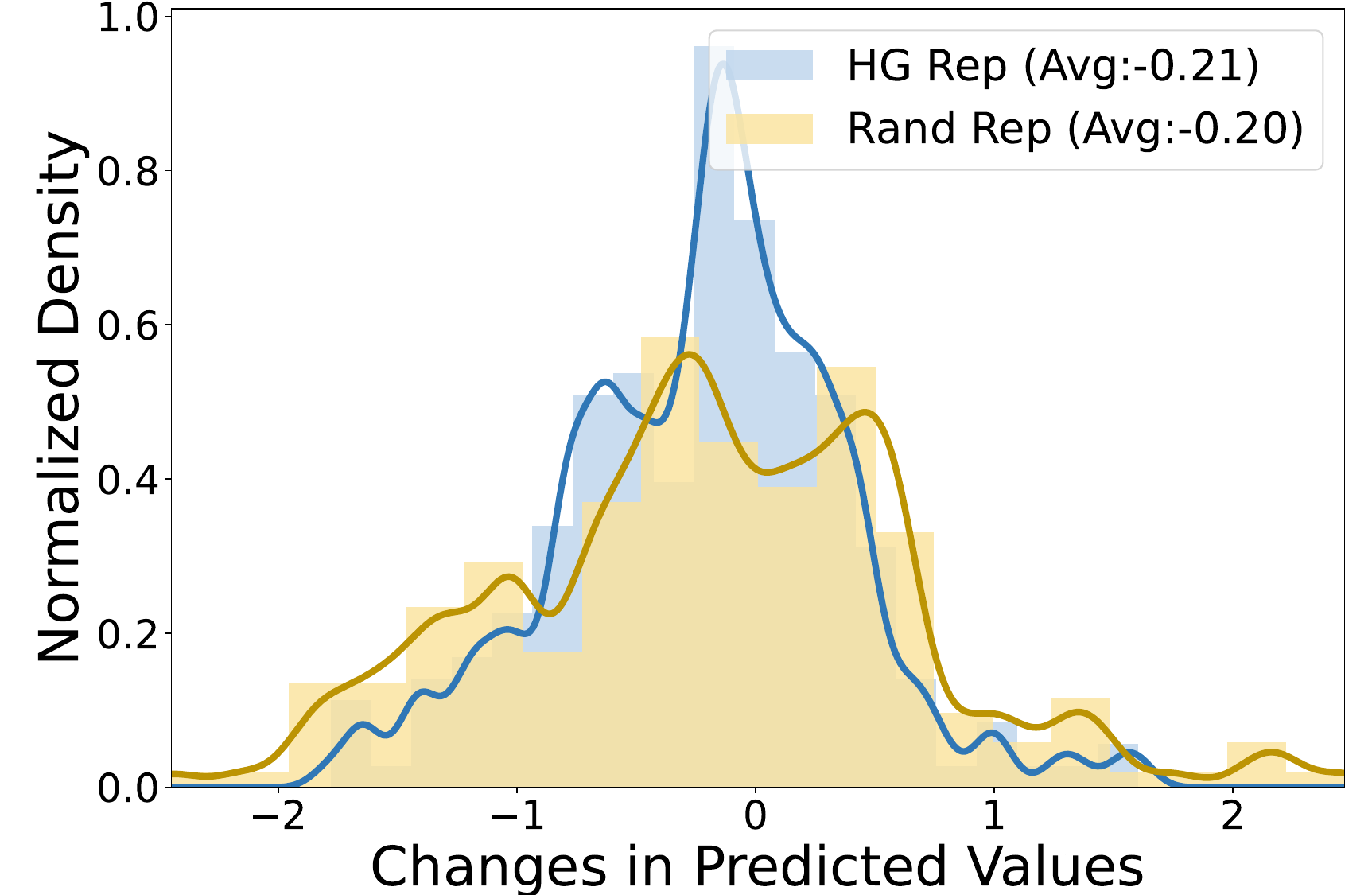}
  }

  \caption{\textbf{Substructure Semantics Modeling on ESOL Dataset.} We compared the effects of two molecular perturbation methods on the \method{}'s and \mlmmth's predictions of hydrophilicity. Figure \ref{fig:substruct_rep_subfig1} show that the impact of replacing hydrophilic groups (HG Rep) and randomly replacing atoms (Rand Rep) on the model's predictions differs significantly, both in the average change in prediction values and their distributions. }    
  \label{fig:substruct_esol_rep}            
\end{figure}

{To eliminate the influence of molecular weight, we designed a hydrophilic group replacement scheme (HG Rep).} We replaced all hydrophilic groups in a molecule with non-hydrophilic groups of similar molecular weight (e.g., methyl, ethyl, propyl) and compared this hydrophilic group replacement scheme (HG Rep) with a random group replacement scheme (Rand Rep), where random groups were replaced with others of similar weight. The results, shown in Figure \ref{fig:substruct_esol_rep}, reveal that \method{} effectively distinguishes between HG Rep and Rand Rep, demonstrating its ability to model key molecular group semantics. It also correctly predicts that replacing hydrophilic groups reduces molecular solubility.

\begin{figure*}[t]

\centering
	\includegraphics[width=0.6\linewidth]{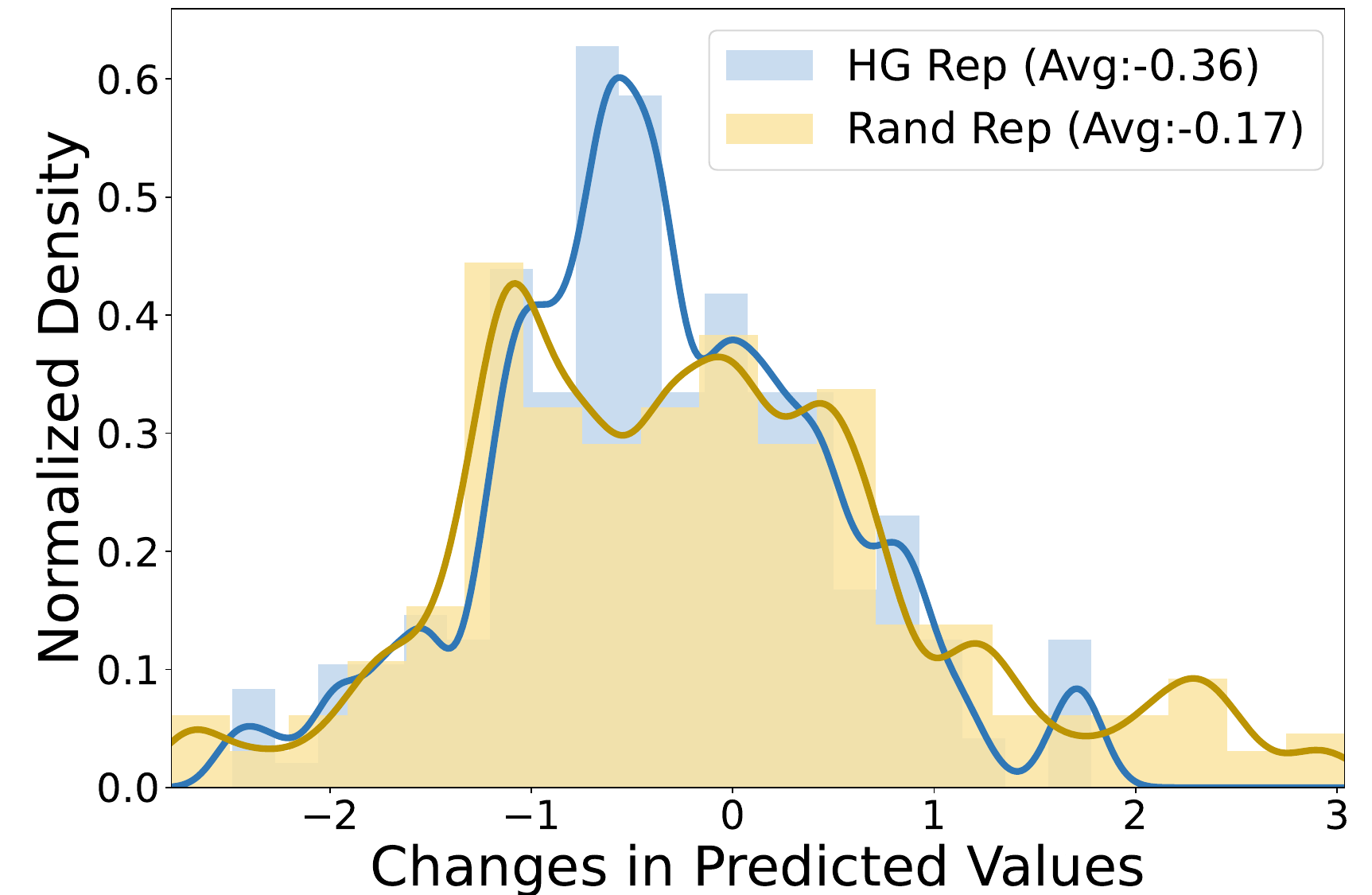}
 
	\caption {Substructure Semantics Modeling on ESOL Dataset of Auto-regressive LM.}
	\label{fig::substruct_ar}
\end{figure*}

Furthermore, we plotted the distribution of predicted changes for MLMs and Auto-regressive LMs before and after these replacement operations in Figure \ref{fig::substruct_ar}. The results show that these models perform significantly worse than the \method{} in distinguishing between random replacements and hydrophilic group replacements. This further highlights the superiority of the \method{} in modeling the semantics of molecular substructures.

\section{\method{}'s Performance on Retrosynthesis Prediction Tasks}
\label{sec::retrosynthesis}

Considering that the original \method{} is an encoder-only model and cannot be directly applied to generative tasks, we further pretrained a model based on an encoder-decoder architecture, referred to as \textbf{\method{}-Gen}. We tested its performance on the retrosynthesis prediction task, where it achieved state-of-the-art results. Below is a detailed discussion:

\textbf{Model Details of \method{}-Gen.} \method{}-Gen adopts a transformer architecture with a base-sized scale \citep{vaswani2017attention}. During pretraining, the input to the encoder consists of SMILES strings with missing molecular fragments, while the decoder’s pretraining task is to reconstruct the original SMILES. Following approaches commonly used in machine translation \citep{vaswani2017attention} , the features extracted by the encoder are passed to the decoder through encoder-decoder attention \citep{vaswani2017attention} . Compared to \method{}, the most significant difference is that the encoder-decoder architecture enables \method{}-Gen to perform sequence-to-sequence generative tasks, allowing us to explore the model’s capabilities in such tasks.

\textbf{ \method{}-Gen Exhibits Strong Performance in Retrosynthesis Prediction Tasks.} Following the experimental setup of EditRetro\citep{han2024retrosynthesis}, we evaluated \method{}-Gen on the retrosynthesis prediction task. During fine-tuning, we applied the same fine-tuning strategies and data augmentation techniques as EditRetro. The experimental results, shown in Table \ref{tab:topk_accuracy}, demonstrate that \method{}-Gen achieved strong performance on the USPTO-50K dataset. This validates that the pretraining approach proposed by \method{} also exhibits excellent performance and great potential in generative tasks.

\begin{table}[htbp]
\centering
\caption{Top-$k$ exact match accuracy of \method{} and baselines on the USPTO-50k dataset.}
\label{tab:topk_accuracy}

\begin{tabular}{lcccc}
\toprule
             & Top-1~$\uparrow$ & Top-3~$\uparrow$ & Top-5~$\uparrow$ & Top-10~$\uparrow$ \\
\midrule
RetroPrime   & 51.4\%     & 70.8\%     & 74.0\%     & 76.1\%     \\
Transformer  & 42.4\%     & 58.6\%     & 63.8\%     & 67.7\%     \\
SCROP        & 43.7\%     & 60.0\%     & 65.2\%     & 68.7\%     \\
MEGAN        & 48.1\%     & 70.7\%     & 78.4\%     & 86.1\%     \\
GTA          & 51.1\%     & 67.6\%     & 74.8\%     & 81.6\%     \\
Retroformer  & 53.2\%     & 71.1\%     & 76.6\%     & 82.1\%     \\
Graph2Edits  & 55.1\%     & 77.3\%     & 83.4\%     & 89.4\%     \\
R-SMILE      & 56.3\%     & 79.2\%     & 86.2\%     & \textbf{91.0\%}     \\
EditRetro    & 60.8\%     & 80.6\%     & 86.0\%     & 90.3\%     \\
\method{}   & \textbf{61.2\%} & \textbf{80.9\%} & \textbf{86.4\%} & 89.7\%     \\
\bottomrule
\end{tabular}

\end{table}

\section{SPE Tokenizer Does Not Improve SMILES MLM Performance}
\label{sec::spe}

We trained a SMILES MLM with SPE tokenizer, \textbf{\mlmmth(SPE)}, using the same architecture and hyperparameters as \method{}, and evaluated it on multiple tasks. As shown in Table \ref{tab:mlm_performance}, \mlmmth(SPE) performs similarly to \mlmmth and significantly worse than \method{}. This demonstrates that introducing SPE cannot replicate the effectiveness of \method{}. The reasons are as follows:
\begin{itemize}
    \item \textbf{Limited Fragment Diversity}: SPE relies on a fixed vocabulary, limiting the diversity of fragment-level information it can capture. In contrast, \method{} dynamically fragments molecules using the BRICS algorithm, capturing a wider variety of molecular substructures.
    \item \textbf{Topology Information Leakage}: SPE-based models retain token position information, which is tied to molecular topology in SMILES, making the prediction task easier but less effective.
    \item \textbf{Lack of Chemical Context}: \method{} fragments molecules based on chemical rules, allowing it to capture substructure information more relevant to molecular properties, unlike SPE, which relies on character pair frequencies.
    \item \textbf{Superior Performance with Fragment-Level Supervision}: A MLM trained with fragment-level supervision, \mlmmth(Frag), outperforms \mlmmth(SPE), as shown in Table \ref{tab:mlm_performance}. This validates the effectiveness of \method{}’s training approach.
\end{itemize}

\begin{table}[htbp]
\centering
\caption{Performance comparison of MLMs with different pretraining strategies.}
\label{tab:mlm_performance}
\begin{tabular}{lccccc|c}
\toprule
               & BACE$\uparrow$ & BBBP$\uparrow$ & SIDER$\uparrow$ & Tox21$\uparrow$ & ToxCast$\uparrow$ & Mean$\uparrow$ \\
\midrule
\mlmmth        & 77.8           & 68.6           & 61.2            & 75.1            & 64.9              & 69.5           \\
\mlmmth(SPE)   & 76.7           & 71.1           & 59.3            & 74.7            & 65.3              & 69.4           \\
\mlmmth(SPAN)  & 78.6           & 67.2           & 59.4            & 76.1            & 62.3              & 68.7           \\
\mlmmth(Frag)  & 79.4           & 73.3           & 62.1            & 74.0            & 64.8              & 70.7           \\
\method{}     & \textbf{80.3}  & \textbf{77.4}  & \textbf{63.0}   & \textbf{77.1}   & \textbf{67.4}     & \textbf{73.0}  \\
\bottomrule
\end{tabular}
\end{table}

\section{Masked Span LM Does Not Improve SMILES LM Performance}
\label{sec::MSLM}

To highlight the differences between \method{} and Masked Span LMs (MSLMs), we trained a SMILES model using MSLM, which randomly masks continuous sequences in SMILES and predicts the missing parts (similar to SpanBERT\citep{joshi2020spanbert}). This model, referred to as \textbf{\mlmmth(SPAN)}, shows performance comparable to \mlmmth but significantly worse than \method{} (see Table \ref{tab:mlm_performance}). This further demonstrates \method{}'s advantages over traditional MSLMs.
Reasons for Poor Performance of Traditional MSLMs:
\begin{itemize}
    \item \textbf{Differences between Text Data and SMILES Data.} Unlike text, molecular data has complex topological structures. In text, adjacent tokens often have strong semantic relevance, and continuous spans convey related information, making span masking effective for learning local semantics. However, \textbf{SMILES lacks such locality}; a single functional group may not appear contiguous, and adjacent tokens may lack strong relevance. For example, aromatic rings with multiple substituents often appear discontinuous in SMILES (we provide a specific case \textbf{CASE1}). This limits the effectiveness of applying span masking directly to SMILES data.
    \item \textbf{Traditional MSLM (e.g., T5\citep{raffel2020exploring}) and \method{} Have Different Implementations; Traditional MSLM is Unsuitable for SMILES Data.} Text data's semantic continuity enables models like T5 to use random span masking, where continuous text segments are masked for prediction. In contrast, SMILES lacks this continuity, so \method{} uses a fragmentation algorithm to split molecules into chemically meaningful fragments. The model predicts missing fragments, which may not correspond to continuous SMILES segments. Unlike traditional MSLM, \method{} focuses on masking chemically significant fragments, a key difference in its design.
    \item \textbf{Better Performance of \mlmmth(Frag).}  The improved performance of \mlmmth(Frag) over \mlmmth(SPAN) highlights the superiority of \method{}’s fragment-level supervision. While \mlmmth(SPAN) uses the traditional MSLM approach, \mlmmth(Frag) incorporates supervision signals similar to \method{}, enabling it to better capture molecular substructure information.
\end{itemize}

\textbf{CASE1} When does SMILES exhibit discontinuity: SMILES is a linearized representation of graph-structured molecules, which inherently causes discrepancies between molecular topology and sequence-level representation. For example, when a ring contains multiple substituents, its representation in SMILES often becomes discontinuous. Consider Glibenclamide, a drug used for diabetes treatment, with the canonical SMILES:  COc1ccc(Cl)cc1C(=O)NCC\textbf{c2ccc}(S(=O)(=O)NC(=O)NC3CCCCC3)\textbf{cc2}. Here, the bolded atoms originate from the same aromatic ring, but due to the multiple substituents, this ring is represented discontinuously in SMILES. Additionally, the aromatic carbon \textbf{cc2} is adjacent to CCC3 atoms from a distant cycloalkane ring. Such discontinuities are common in SMILES and adversely affect Masked Span LMs.

\section{Comparison between SMI-Editor and contrastive learning}
\label{sec::CL}

\textbf{Similarities}: Both contrastive learning and SMI-Editor aim to learn alignment.
\begin{itemize}
    \item \textbf{Contrastive learning aligns representations of different views.} The core idea of contrastive learning is to bring the representations of different views of the same sample (positive pairs) closer while pushing representations of different samples (negative pairs) apart. Essentially, this process learns the correct alignment between views of the same sample.  
    \item \textbf{SMI-Editor aligns representations of missing substructures and contexts.} As \citet{fu2022contextual} noted, MLMs align the representations of contexts and missing words during training. Similarly, SMI-Editor aligns the representations of missing substructures and their contexts. For example, given the input Nc1ccc(O)cc1, the model need to predict the complete molecule CC(=O)Nc1ccc(O)cc1. SMI-Editor can effectively align the representation of the missing fragment CC(=O) with the context Nc1ccc(O)cc` through this process.
\end{itemize}

\textbf{Differences}: The alignment targets differ between the two paradigms.  

\begin{itemize}
    \item \textbf{Contrastive learning focuses on global information}: The representations to be aligned often correspond to different augmented views of the same molecule, such as through atom deletion, bond deletion, or subgraph deletion. These views typically preserve the molecule's overall structure and thus contain global information.  
    \item \textbf{SMI-Editor emphasizes aligning local substructure information with global context}: In SMI-Editor, the context typically corresponds to the molecule's backbone, representing global information, while the missing substructures contain local information.  
    \item \textbf{SMI-Editor is more sensitive to local structure information}: By aligning local substructures with global context, SMI-Editor learns finer-grained semantics from SMILES data, making it better suited to capturing detailed molecular information than contrastive learning.  
\end{itemize}

\section{K-fold cross-validation of the \method{} model.}
\label{sec::k_fold}

Using a $5$-fold setup, we evaluated \method{}’s performance on the training sets of BACE, BBBP, SIDER, Tox21, ToxCast, ClinTox and MUV. The results are shown in Table \ref{tab:cross_validation_results}. These results demonstrate that \method{} exhibits strong performance and stability across downstream tasks.

Each dataset was evenly divided into five parts. In each run, one part was selected as the validation set, while the remaining four parts were used as the training set. The model was trained and evaluated on the validation set. This process was repeated five times to complete all runs.

\begin{table}[htbp]
\centering
\caption{5-fold cross-validation results of the \method{} model.}
\label{tab:cross_validation_results}
\begin{tabular}{lccccccc}
\toprule
          & BACE$\uparrow$ & BBBP$\uparrow$ & SIDER$\uparrow$ & Tox21$\uparrow$ & ToxCast$\uparrow$ & ClinTox$\uparrow$ & MUV$\uparrow$\\
\midrule
Run 1     & 91.92 & 97.64 & 62.59 & 83.69 & 75.83 & 99.76 & 77.23 \\
Run 2     & 91.86 & 96.27 & 66.89 & 84.09 & 73.31 & 98.73 & 79.49 \\
Run 3     & 90.82 & 98.53 & 62.60 & 84.87 & 73.52 & 99.6 & 77.4 \\
Run 4     & 91.13 & 98.77 & 63.32 & 83.95 & 74.60 & 99.82 & 77.58 \\
Run 5     & 90.68 & 97.84 & 63.50 & 85.83 & 75.51 & 98.61 & 79.39 \\
\midrule
\textbf{Mean} & \textbf{91.28} & \textbf{97.81} & \textbf{63.78} & \textbf{84.48} & \textbf{74.55} & \textbf{99.30} & \textbf{78.23} \\
\textbf{Std}  & \textbf{0.58}  & \textbf{0.97}  & \textbf{1.78}  & \textbf{0.87}  & \textbf{1.13} & \textbf{0.59} & \textbf{1.12} \\
\bottomrule
\end{tabular}
\end{table}

\section{Broad Applications of Atom-Level Tokenizers}
\label{sec::atom_tok}

Currently, many SMILES LMs, including MLM and autoregressive LMs, rely on atom-level tokenizers to process molecular representations. Atom-level tokenizers break down SMILES strings into individual atomic units or tokens, such as atoms and simple symbols (e.g., ``C'', ``O'', ``=''). This approach simplifies the tokenization process and aligns well with the intrinsic atomic structure of molecules, enabling models to capture fine-grained atomic interactions and features. For example, MolXPT \citep{liu2023molxpt} and Dual-view Molecular Pre-training \citep{zhu2023dualview} explicitly leverage atom-level tokenization to enhance the granularity of molecular representations, facilitating downstream tasks such as molecule generation and property prediction.

Atom-level tokenization has the advantage of maintaining a straightforward correspondence between the SMILES representation and the underlying molecular structure, making it easier for the model to interpret local chemical environments. This granularity is particularly beneficial for tasks that require precise predictions. For instance, studies such as ChemBERTa \citep{chithrananda2020chemberta}, Molecular Transformer \citep{schwaller2019molecular}, and SMILES-BERT \citep{wang2019smilesbert} demonstrate that atom-level tokenization can achieve good performance in molecular property prediction tasks.

}
\end{document}